\documentclass[10pt,journal,compsoc]{IEEEtran}
\usepackage{graphicx,amsmath,amssymb,shortbold,verbatim,multirow,booktabs,colortbl,wrapfig}
\definecolor{myred}{rgb}{0.5,0,0}
\definecolor{mygreen}{rgb}{0,0.5,0}
\definecolor{myblue}{rgb}{0,0,0.5}

\ifCLASSOPTIONcompsoc
  \usepackage[nocompress]{cite}
\else
  \usepackage{cite}
\fi
\begin{document}
\title{From Images to 3D Shape Attributes}
\author{David~F.~Fouhey, Abhinav~Gupta, Andrew~Zisserman
\IEEEcompsocitemizethanks{\IEEEcompsocthanksitem D.F. Fouhey is with the EECS
Department, University of California Berkeley, Berkeley, CA 94720
\IEEEcompsocthanksitem A. Gupta is with the Robotics Institute, Carnegie Mellon
University, Pittsburgh, PA 15232
\IEEEcompsocthanksitem A. Zisserman is with the Visual Geometry Group, Department of Engineering
Science, University of Oxford, Oxford, OX1 3PJ, UK}}

\markboth{To Appear in TPAMI}
{Shell \MakeLowercase{\textit{et al.}}: Bare Demo of IEEEtran.cls for Computer Society Journals}

\IEEEtitleabstractindextext{
\begin{abstract}
Our goal in this paper is to investigate properties of 3D shape that
can be determined from a single image.  We define {\em 3D shape
attributes} -- generic properties of the shape that capture curvature,
contact and occupied space. Our first objective is to infer these 3D
shape attributes from a single image. A second objective is to infer a
{\bf 3D shape embedding} -- a low dimensional vector representing the
3D shape.

We study how the 3D shape attributes and embedding can be obtained
from a single image by training a Convolutional Neural Network (CNN)
for this  task.  We start with synthetic images so that the contribution
of various cues and nuisance parameters can be controlled. 
We then turn to real images and introduce a large scale image dataset of
sculptures containing 143K images covering 2197 works from 242 artists.

For the CNN trained on the sculpture dataset we show the following:
(i) which regions of the imaged  sculpture  are used by the  CNN to infer the 3D shape attributes; (ii)
that the shape embedding  can be used to match
previously unseen sculptures largely independent of viewpoint; and (iii)
that the 3D attributes 
generalize to images of other
(non-sculpture) object classes.
\end{abstract}

}

\maketitle

\IEEEdisplaynontitleabstractindextext

\IEEEpeerreviewmaketitle

\section{Introduction}
\label{sec:introduction}

Consider a few objects in your surroundings -- perhaps a cup, a banana, or a
far-off abstract sculpture you see through your window. How might you represent their
3D shape?  In the early days of computer vision, a
menagerie of representations were used to answer this question, each with a particular
niche and relative advantages. However, with a number of notable exceptions,
the field has increasingly turned this grand challenge into the task of
figuring out what a depth sensor might see if it were pointed at the scene,
i.e., a per-pixel metric map.

This paper takes an alternate view and proposes to infer high-level descriptions or generic properties
of shape directly from an image.
We term these properties {\bf 3D shape attributes} and introduce a variety of specific examples,
for instance planarity, thinness, point-contact, to concretely explore this
concept. These shape attributes are a subset of {\bf higher order shape properties}, or properties that go beyond depth, the $0\textsuperscript{th}$ derivative.
Other examples include normals and curvature. Such properties can, in principle, be derived from an estimated depthmap.
However, in practice, deriving these properties from a depth map is inferior to other methods used by
both humans and  machines for a myriad of reasons, including:
insufficient resolution, view dependence, compounding errors, and the existence of direct
cues for higher order shape properties. We demonstrate this empirically with baselines on
our particular problem as well as summarizing and discussing the evidence from human
perception studies.

As with classical object attributes and relative attributes~\cite{Ferrari07,Farhadi09,Parikh11}, 3D
attributes offer a means of describing 3D object shape when confronted
with something entirely new -- the {\em open world problem}. This is in
contrast to a long line of work which is able to say something about
3D shape, or indeed recover it, from single images
{\em given} a specific object class, e.g.\ faces~\cite{Blanz99},
semantic category \cite{Kar15} or cuboidal room structure
\cite{Hedau09}. While there has been success in determining {\it how}
to apply these constraints, the problem of {\it which} constraints to apply 
is much less explored, especially in the case of
completely new objects. Used inappropriately, scene understanding methods
tend to produce either unconstrained results \cite{Eigen14,Fouhey13a} in which
walls that should be flat bend arbitrarily or planar interpretations
\cite{Fouhey14c,Lee09} in which non-planar objects like lamps are
flat.  Shape attributes can act as a generic way of representing
top-down properties for 3D understanding, sharing with classical
attributes the advantage of both learning and application across
multiple object classes.

There are two natural questions to ask: what 3D attributes
should be inferred, and how to infer them? After further
motivating the problem of studying higher order shape properties in Section~\ref{sec:motivation},
we introduce our attribute vocabulary in Section~\ref{sec:attributes},
which draws inspiration from and revisits past work in both the 
computer and the human vision literature. We return to these ideas with modern
computer vision tools. In particular, as we describe in Section~\ref{sec:approach},
we use Convolutional Neural Networks (CNNs) to infer the 3D attributes from
an image.

A secondary objective of this paper is to obtain a {\bf 3D shape embedding}
-- a low dimensional vector representing the 3D shape of the object.
Again, this is inferred from an image using a CNN, 
and described in Section~\ref{sec:approach}. Our aspiration is that the
embedding should be largely unaffected by the viewpoint of the image.

The next important question is: what data to use to investigate these
properties? We use photos of modern sculptures from Flickr, and
describe a procedure for gathering a large and diverse dataset in
Section \ref{sec:dataset}.  This data has many desirable properties:
it has much greater variety in terms of shape compared to common-place
objects; it is real and in the wild, so has all the challenging
artifacts such as severe lighting and varying texture that may be
missing in synthetic data. Additionally, the dataset is automatically
organized into: {\it artists}, which lets us define a train/test
split to generalize over artists; {\it works} (of art) irrespective of
material or location, which lets us concentrate on shape, and {\it
viewpoint clusters}, which lets us recognize sculptures from multiple
views and aspects.

Our experiments show that we are indeed able
to infer 3D shape attributes. We begin by verifying our network with a series of control experiments
akin to psychophysics in Section \ref{sec:analysissynthesis}. We subsequently
analyze the network on our dataset of sculpture in Section \ref{sec:experiments}.  
However, we also ask the question of whether we are
actually learning 3D properties, or instead a proxy property, such as the
identity of the artist, which in turn enables these properties to be inferred.
We have designed the experiments both to avoid this possibility and to probe
this issue, and discuss this there. 

This paper is an extension of our previous work~\cite{Fouhey16}.
The extensions include: (i) additional motivation for our study
of higher order shape properties as ends themselves in Section \ref{sec:motivation};
(ii) additional description details throughout 
the paper; 
(iii) experiments with {\em synthetic stimuli} in Section \ref{sec:analysissynthesis} that
provide additional validation that the method is learning about 3D properties,
and offer insights into how it uses a mix of shading, contours, and texture; 
(iv) more thorough evaluation of the results, 
such as saliency maps in Section~\ref{sec:saliency}, and failure modes of 
the mental rotation task in Section~\ref{sec:exp_rot}.

\section{Related Work}

How do 2D images convey 3D properties of objects?  This is one of the
central questions in any discipline involving perception -- from
visual psychophysics to computer vision to art. Our approach draws on
each of these fields, for instance in picking the
particular attributes we investigate or probing our learned model.

One motivation for our investigation of shape attributes is a long
history of work in the human perception community that aims to go
beyond metric properties and address holistic shape in a
view-independent way. Amongst many others, Koenderink and van Doorn
\cite{Koenderink92} argued for a set of shape classes based on the
{\it sign} of the principal curvatures and also that shape
perception was not metric~\cite{koenderink1995relief,koenderink92surface}, 
and Biederman
\cite{Biederman87} advocated shape classes based on non-accidental
{\it qualitative} contour properties. 

We are also inspired by work on trying to use mental rotation
\cite{Shepard71,Tarr1998} to probe how humans represent shape; here,
we use it to probe whether our models have learned something sensible.
A great deal of research in early computer vision sought to extract
local or qualitative cues to shape, for instance from apparent contours~\cite{Koenderink84},
self-shadows and specularities~\cite{Koenderink90,Zisserman89}.
Recent computer vision
approaches to this problem, however, have increasingly reduced 3D understanding
to the task of inferring a viewpoint-dependent 3D depth or normal 
at each pixel \cite{Barron15,Eigen14,Fouhey13a}, with most recent works
developing the idea of
inferring a point-set or voxel-based 
 3D shape given a set of classes (e.g.\ cars, chairs, rooms)
and a large dataset of synthetic 3D models of those classes for
training~\cite{Fan16,Girdhar16b,Tulsiani17,Wu16,Yan16}. 
These predictions are useful
for many tasks but do not tell the whole story, as we argued in the introduction.
This work aims to help fill this gap by revisiting these non-metric 
qualitative questions.
Some exceptions to this trend include the qualitative labels explored in \cite{Hoiem05,Gupta10}
like porous, but these initial efforts had limited scope in terms of data variety,
vocabulary size, and quantity of images.

Our focus is 3D shape understanding, but we pose our investigation into these
properties in the language of attributes
\cite{Farhadi09,Ferrari07,Parikh11,Kumar09,Lampert09c} to emphasize their key
properties of communicability and open-world generalization.  The vast majority
of attributes, however, have been semantic and there has never been, to our
knowledge, a systematic attempt to connect attributes with 3D understanding or
to study them with data specialized for 3D understanding. Our work is most
related to the handful of coarse 3D properties in \cite{Farhadi09} or
the 3D shape properties extracted from 3D models proposed in \cite{Gong13}. 
Compared to \cite{Farhadi09}, in addition to having a larger number of shape attributes and data designed for
3D understanding, our attributes are largely unaffected by viewpoint change. 
In contrast to \cite{Gong13}, our work focuses on the complementary
problem of perception in images as opposed to 3D models and exclusively on 
shape properties as opposed to functional ones.

\section{Directly Modeling Higher-Order Properties of Shapes}
\label{sec:motivation}

In this paper, we study higher-order shape properties.
These are properties of shapes that are not simple depthmaps.

Surface normals are the simplest example, but there are many other properties
that have received far less attention in the literature:
we investigate some of these, including 
planarity, roughness, and topological genus. 

Why should we study higher-order properties of shape as entities
in themselves, and
not as the result of analyzing a property like a depthmap? In principle,
with sufficient resolution and accuracy, a depthmap contains all
the information necessary to construct many higher order properties: 
the normals and curvatures by taking first and second derivatives,
and many others by applying the right analyses.
It is thus possible that by obtaining a depthmap,
one should get higher-order properties for free via this {\it indirect} method.
While simple, the indirect approach is contradicted by evidence from both
humans and machines. 

Evidence in psychophysics suggests that the human visual system
employs multiple types of representations of shape, and that 
some properties, which in principle could be derived from depth,
are instead obtained {\em directly}. Both Koenderink {\it et al.}~\cite{Koenderink96} and Norman
and Todd \cite{Norman96} found that the accuracy of orientation estimates
could be substantially higher than differentiating 
estimated depth ought to permit. Johnston and Passmore 
\cite{Johnston94} found similar results with orientation and curvature. 

The human results of Koenderink {\it et al.} and Norman and Todd can be reproduced in
machines. Consider the recent approach of \cite{Eigen15} that predicts both depth
and surface normals from an image with an identical CNN architecture. 
We can compare the indirect method of computing normals from estimated
depth to the direct method of estimating normals using the standard NYUv2 
dataset \cite{Silberman12} and the ground-truth of \cite{Ladicky14}. 
While the indirect normals are reasonable, the accuracy still
lags far behind the direct method ($30.3^\circ$ vs $20.9^\circ$
mean error)

and is worse in all normal
metrics of~\cite{Fouhey13a} (and it should be noted that this error gap is
probably a best case, since the depth loss of~\cite{Eigen15} already
incorporates a local normal term). 

Why might it be the case that the seemingly straightforward 
notion of obtaining higher-order properties for free indirectly
does not lead to good estimates in practice? In addition to pitfalls of all
indirect modeling, such as irreversible error accumulation, we 
outline a few reasons below:

\begin{figure}
\includegraphics[width=\linewidth]{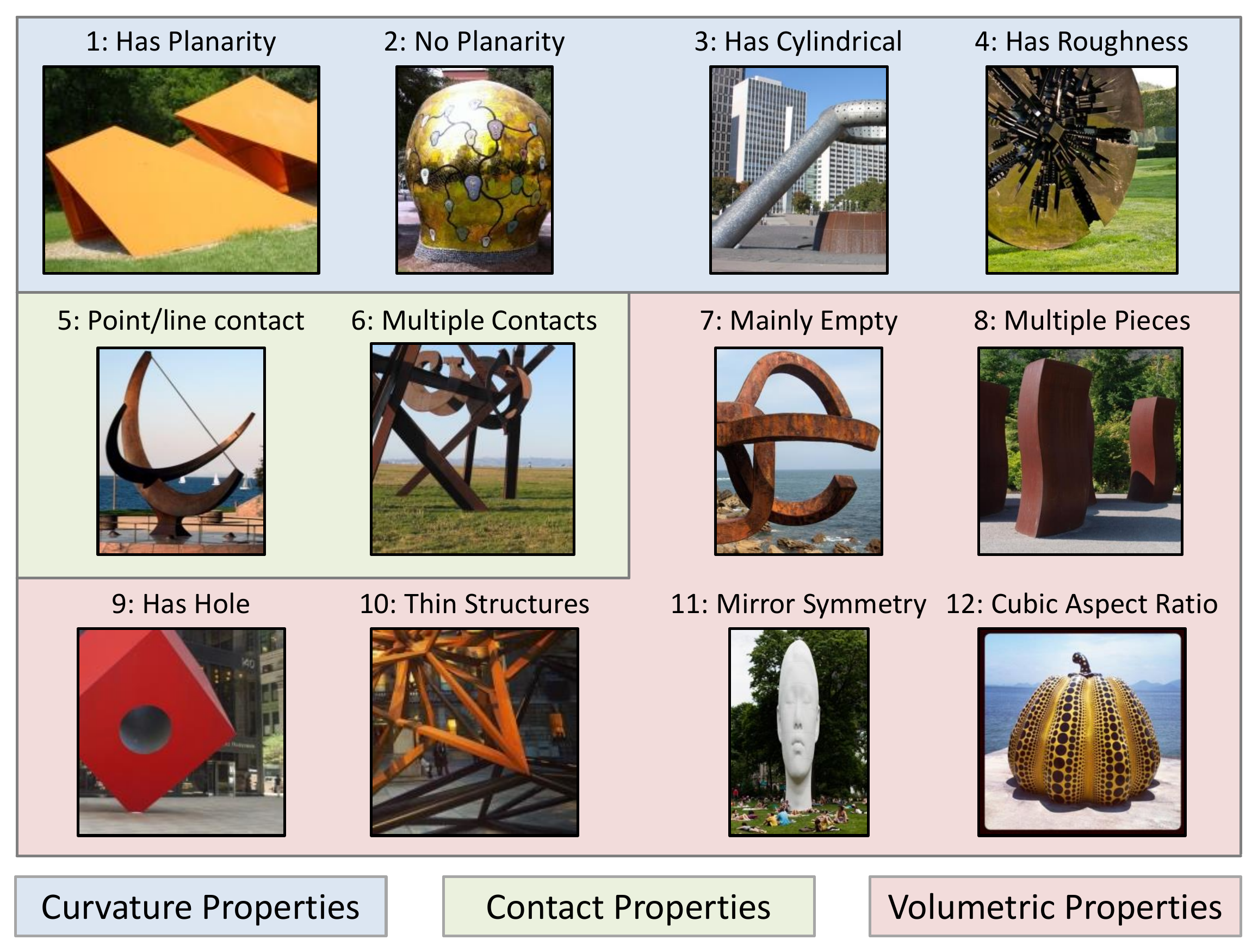} 
\caption{The 3D shape attributes investigated in this paper, and
an illustration of each from our training set. Additional
sample annotations are shown in Fig.~\ref{fig:attrExample}.}
\label{fig:attrTable}
\end{figure}

\noindent {\bf Direct cues for higher order properties:} 
One argument in favor of the direct approach is that many ``cues for depth'',
are actually direct cues for higher order properties, and thus converting them 
first into
cues for depth is suboptimal. Examples include texture gradients
\cite{Gibson50,Cutting95}, which convey changes in surface 
orientation~\cite{Forsyth01}, or 
the curvature of occluding contours \cite{Koenderink84}, which indicates
the sign of the Gaussian curvature of the shape.

\noindent {\bf Resolution:} Consider determining
if a wire fence has thin structures or a piece of
sandpaper has a rough surface. Compared to simply recognizing
wires and bumps, the indirect method requires
interpreting the scene at an incredibly detailed resolution
-- high enough to capture the pixels of the fence wire and sub-millimeter bumps on the
sandpaper.

\noindent {\bf Ambiguity:} Finally, ambiguities in depth may not be ambiguities
for higher order properties, and prematurely resolving them in terms of
depth is often the wrong thing to do. 
For instance, consider observing a surface and having 
three plausible hypotheses for its shape: convex ($z=x^2$), concave ($z=-x^2$), and flat ($z=0$).
Suppose one is overwhelmingly confident ($>95\%$ chance) it is not flat but places equal chance
on it being the other possibilities. Even though the surface is unambiguously {\it not} flat,
the correct surface with regards to depth in both the $L_1$- and $L_2$-norm
sense is a flat surface. If the ambiguity is resolved in depth, the resulting
interpretation in terms of higher order properties is radically and incorrectly
altered. Instead, if one directly asks whether the curvature is non-zero, the correct
answer is obtained.

\section{3D Attribute Vocabulary}
\label{sec:attributes}

Which 3D shape attributes should we model?  We choose 12 attributes
based on questions about three properties of historical interest to
the vision community -- curvature (how does the surface curve locally
and globally?), ground contact (how does the shape touch the ground?),
and volumetric occupancy (how does the shape take up space?).

\begin{figure}
\includegraphics[width=\linewidth]{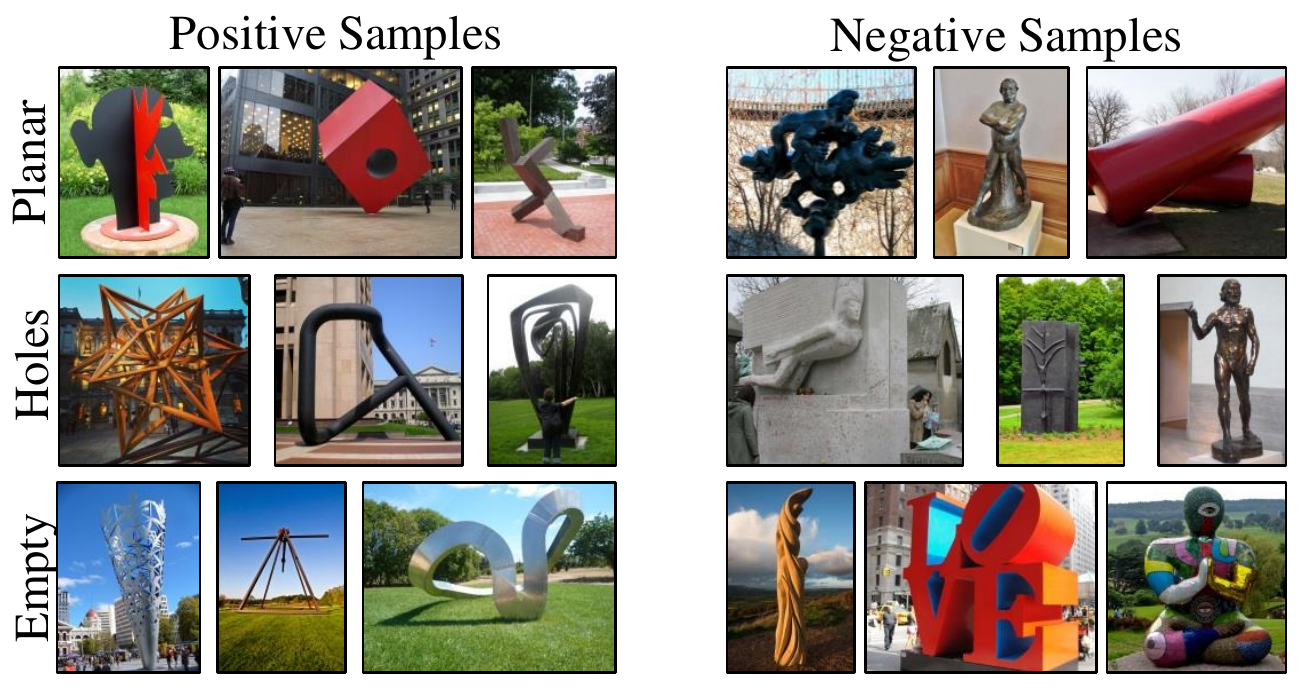}
\caption{Sample positive and negative annotations from the dataset 
for the planar, has-holes, and empty attributes. }
\label{fig:attrExample}
\end{figure}

Fig.~\ref{fig:attrTable} illustrates the 12 attributes, and 
sample annotations are shown in Fig.~\ref{fig:attrExample}. We now briefly describe the attributes in terms of
curvature, contact, and volumetric occupancy. \\
\noindent {\bf Curvature Attributes:} We take inspiration
from a line of work on shape categorization via curvature led by Koenderink and van Doorn (e.g.,
\cite{Koenderink92}). Most sculptures have a mix of convex, concave, and saddle
regions, so we analyze where curvature is zero in at least one direction and look for (1) {\it Has Planarity:} piecewise planar sculptures;
(2) {\it Has No Planarity:} sculptures with {\it no} planar regions (note that many sculptures have a mix of planar and non-planar regions); (3) 
{\it Has Cylindrical:} sculptures where one principal curvature is zero (e.g., cylindrical ones); and 
(4) {\it Has Roughness:} rough sculptures where locally the surface changes rapidly. \\
\noindent {\bf Contact Attributes:} Contact and support reasoning plays a strong role in scene understanding
(e.g., \cite{Hoiem08,Gupta10,Hedau12,Silberman12,Guo13}). We characterize ground contact via
(5) {\it Point/line Contact:} point or line contact as compared to contact with the full body;  
(6) {\it Multiple Contacts:} whether multiple contacts between the object and the ground are made.\\
\noindent {\bf Volumetric Attributes:} Reasoning about
occupied-space has long been a goal of 3D understanding \cite{Hedau12,Lee10,Rock15}. We ask
(7) {\it Mainly Empty:} the fraction of occupied space in the sculpture; 
(8) {\it Multiple Pieces:} whether the sculpture has multiple pieces; 
(9) {\it Has Hole:} whether there are holes (i.e., the topology of the sculpture); 
(10) {\it Has Thin Structures:} whether it has thin structures, irrespective of whether they are sheets
or tubular; (11) {\it Mirror Symmetry:} whether it is reflection symmetric i.e., if there is a plane of mirror symmetry in 3D; and 
(12) {\it Cubic Aspect Ratio:} whether it has a cubic aspect ratio in 3D. 

Note that of the 12 attributes, 10 are relatively unaffected by a geometric
affine transformation of the image (or 3D space) -- only the 
mirror symmetry and cubic aspect ratio attributes are measuring a global metric property.

These are, of course, not a complete set. We do not model, for example, enclosure
properties or differentiate a single large hole from a mesh. Similarly, many properties,
such as Koenderink and Van Doorn's shape index or Beiderman's geons are
localized or part-based. We leave this to future work.

\begin{figure*}
\includegraphics[width=\textwidth]{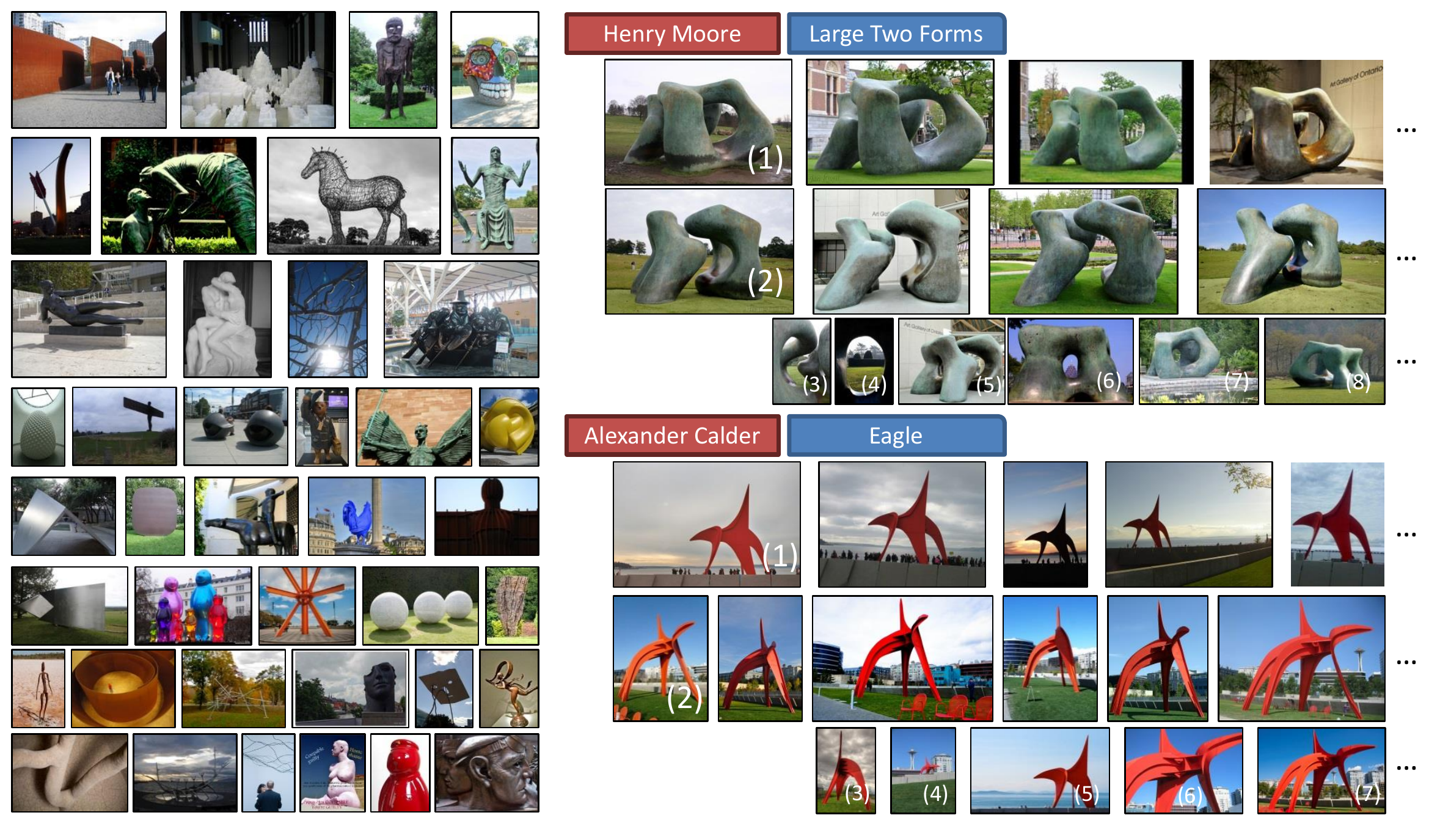}
\caption{The dataset consists of 143K images of sculptures that were
gathered from Flickr and Google images. A representative sample is shown on 
the left. Note the great variety in shape, material, and style. 
Our data has structure in terms of artist, work, and viewpoint cluster (shown numbered  on the right). Each
is important for investigating 3D shape attributes.
}
\label{fig:samples}
\end{figure*}

\section{Gathering a Dataset of 3D Shapes}
\label{sec:dataset}

In order to investigate these 3D attributes, we need a dataset of 3D shapes
that has a diversity of shape so that different subsets of attributes apply.
We use modern sculptures as our source of 3D shapes
since they are diverse in form and in-the-wild photos of them are available in
great quantities on the Internet.

One alternative would be to use ordinary objects, such as the 20 PASCAL objects
\cite{Everingham10}. Unfortunately, ordinary objects have limited diversity,
not just in terms of overall combinations of shape attributes, but also
in terms of shape attributes conditioned on category. In practice, this means
that if we set out to study shape with ordinary objects, our learning models
may simply exploit categories as proxy variables: 
for example, rather than analyze planarity, our models may take the short-cut of distinguishing
people from trains, then predicting planarity accordingly. In contrast, 
sculpture is free to depict people as planar or objects that defy categorization.

While using modern sculpture helps prevent a trivial solution, artists often
produce work in a similar style: Alexander Calder's sculptures are mostly
piecewise planar, Constantin Brancusi's egg-shaped,
and Henry Moore's are smooth and non-planar. We therefore need a
variety of artists and multiple works/images of each.
Previous sculpture datasets~\cite{Arandjelovic11,Arandjelovic12a} are not
suitable for this task as they only contain a small number of artists
and viewpoints.

Thus we gather a new dataset from Flickr. We adopt a five stage process to
semi-automatically do this: (i) obtain a vocabulary of artists and works (for
which many images will be available); (ii) cluster the works by viewpoint;
(iii) clean up mistakes; (iv) query expand for more examples from Google
images; and (v) label attributes.  Note, organization by artist is not strictly
necessary. However, artists are used subsequently to split the works into train
and test datasets: as noted above, due to
an artists' style, shape attributes frequently correlate with an
artist; consequently artists in the train and test splits must be
disjoint to avoid an overly optimistic generalization performance.
The statistics for these stages are given in~Tab.~\ref{tab:stats}.

\subsection{Generating a vocabulary of artists and works}

Our goal is to generate a vocabulary of artists and works that is as
broad as possible. 
We begin by producing a list of artists, combining manually generated
lists with automatic ones, and then expand each artist to a list of their
works.

The manual list consists of the artists
exhibited at six sculpture parks picked from online top-10 lists, as
well as those appearing in Wikipedia's article on Modern Sculpture.
An automatic list is generated from metadata from the 20
largest sculpture groups on Flickr: we analyze image titles for text
indicating that a work is being ascribed to an artist, and take frequent bigrams and
trigrams. The two lists are manually filtered to remove misspellings,
painters and architects, a handful of mistakes, and artists with
fewer than 250 results on Flickr. This yields 
258 artists (95 from the manual list, and 163 from the automatic).

We now find a list of potential works for each artist using both
Wikipedia and text analysis on Flickr. We query the sculptor's page on
Wikipedia, possibly manually disambiguating, and propose any
italicized text in the main body of the article as a possible work. We
also query Flickr for the artists' works (e.g., Tony Smith Sculpture),
and do n-gram analysis in titles and descriptions in front of phrases
indicating attribution to the sculpture (e.g., ``by Tony Smith''). In
both cases, as in~\cite{Quack08}, stop-word lists were 
effective in filtering out noise. While Wikipedia has high precision,
its recall is moderate at best and zero for most artists. Thus
querying Flickr is crucial for obtaining high quality data.
Finally, images are downloaded from Flickr for each work of each artist.

\subsection{Building viewpoint clusters}
\label{sec:clusters}

\begin{table}
\caption{Data statistics at each stage and the trainval/test splits.}

\centering
\small
\begin{tabular}{l@{~~~}c@{~~~}c@{~~~}c@{~~~}c} \toprule
Stage       & Images & Artists & Works & View. Clusters \\ \midrule
Initial     & 352K & 258 & 3412 & -- \\
View Clust. & 213K & 246 & 2277 & 16K \\
Cleaned       & ~97K & 242 & 2197 & ~9K \\ 
Query Exp.  & 143K & 242 & 2197 & ~9K \\ \midrule
Trainval/Test  & 109K/35K & 181/61 & 1655/532 & 7.2K/2.1K
\\ \bottomrule
\end{tabular}
\label{tab:stats}
\end{table}

Images from each work are  partitioned into {\em viewpoint
clusters}. These clusters are image sets that, for example, capture a
different visual aspect of the work (e.g.\ from the front or
side) or are acquired from a particular distance or scale (e.g.\ a
close up). Fig.~\ref{fig:samples} shows example viewpoint clusters for
several works.

There are two principal reasons for obtaining viewpoint clusters: (i) it
enables recognition of a work from different viewpoints to be evaluated; and
(ii) it makes label annotation more efficient as attributes are in general
valid for all images of a cluster. Note, it might be thought that attributes
could be labelled at the work level, but this is not always the case. For example, 
the hole in a Henry Moore sculpture or the ground contact of an Alexander Calder 
sculpture may not be
visible in some viewpoint clusters, so those clusters will be labelled
differently from the rest (i.e., no hole for the former, and unknown for 
the latter).

Clustering proceeds in a standard manner by defining a similarity
matrix between image pairs, and using spectral clustering over the
matrix.  The pairwise similarity measure takes into
account: (i) the number of correspondences (that there are a threshold
number); (ii) the stability of these correspondences (using cyclic
consistency as in~\cite{Zhou15}); and (iii) the viewpoint change
(the rotation and aspect ratio change obtained from an affine
transformation between the images). Computing correspondences requires some
care though since sculptures often do not have texture (and thus SIFT
like detections cannot be used). We follow~\cite{Arandjelovic10} and first obtain a local boundary descriptor
for the sculpture (by foreground-background segmentation and 
MCG \cite{Arbelaez14} edges for the boundaries), and then obtain geometrically consistent
correspondences using an affine fundamental matrix. Finally, a loose
affine transformation is computed from the correspondences (loose
because the sculpture may be non-planar, hence the earlier use of a
fundamental matrix).

In general, this procedure produces clusters with high purity.  The
main failure is when an artist has several visually similar
works (e.g.\ busts) that are confused in the meta-data used
to download them.  We also experimented with using GPS,
but found the tags to be too coarse and noisy to define satisfactory
viewpoint clusters.

\subsection{Data Cleanup}

The above processes are mainly automatic and consequently make some
mistakes.  A number of manual and semi-automatic post-processing steps
are therefore applied to address the main failings. Note, we can
quickly manipulate the dataset via viewpoint clusters as opposed to
handling each and every image individually.

\noindent {\it Cluster filtering:} Each cluster is 
checked manually using three sample images to reject
clearly impure clusters. 

\noindent {\it Regrouping:} 
Some of the automatically generated works are ambiguous due to noisy
meta-data: for instance ``Reclining Figure'' describes a number of 
Henry Moore sculptures. After clustering, these are reassigned to
the correct works.

\noindent {\it Outlier image removal:} A 1-vs-rest SVM is trained 
for each work, using fc7 activations of a CNN~\cite{Krizhevsky12}
pretrained on ImageNet~\cite{ILSVRC15}. Each work's images are
sorted according to the SVM score, and the bottom images 
($\approx 10K$ across all works) flagged
for verification. 

\subsection{Expansion Via Search Engines}

\begin{figure}
\includegraphics[width=\linewidth]{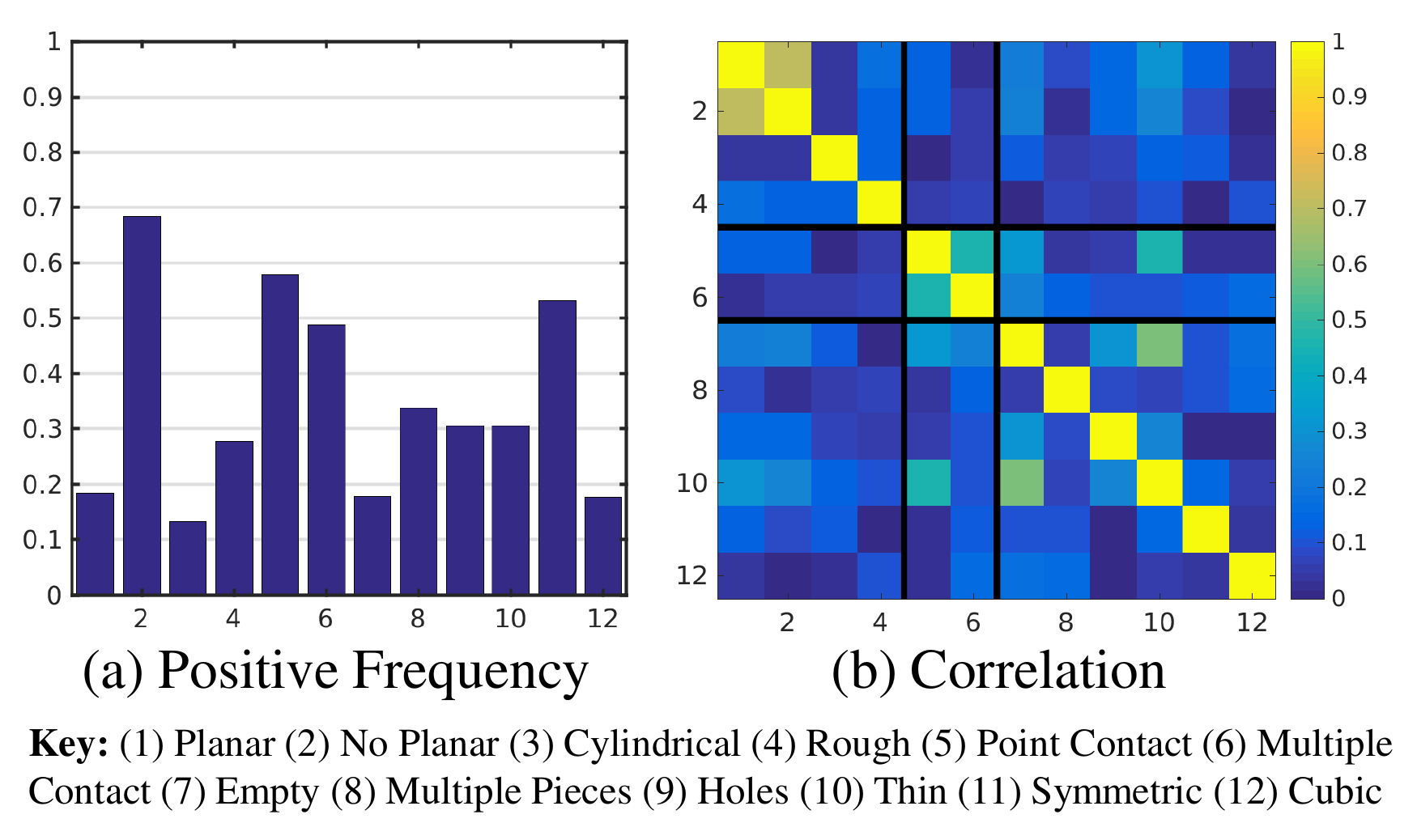}
\caption{(a) Frequency of each attribute (i.e., \# positives /\# labeled);
(b) Correlation between attributes.} 
\label{fig:attrstats}
\end{figure}

Finally, we augment the dataset by querying Google. We perform queries
with the artist and work name. Using the same CNN activation + SVM technique
from the outlier removal stage, we re-sort the query results and add the top
images after verification. This yields $\approx 45K$ more images.

\subsection{Attribute Labeling}

The final step is to label the images with attributes. Here, the viewpoint
clusters are crucial, as they enable the labeling of multiple images at once.
Each viewpoint cluster is labeled with each attribute, or can be labeled as N/A
in case the attribute cannot be determined from the image (e.g., contact
properties for a hanging sculpture). One difficulty is determining a threshold:
few sculptures are only planar and no sculpture is fully empty. We assume an
attribute is satisfied if it is true for a substantial fraction of the
sculpture, typically 80\%. To give a sense of attribute frequency, we show the
fraction of positives in Fig.\ \ref{fig:attrstats}(a).

The dataset is also diverse in terms of 
combinations of attributes and inter-attribute correlation.
There are $2^{12}=4096$ possible combinations, of which 393 occur in our data.
Most attributes are uncorrelated according to the correlation
coefficient $\phi$, as seen in Fig.\ \ref{fig:attrstats}(b):
mean correlation is $\phi = 0.13$
and $82\%$ of pairs have $\phi < 0.2$.
The two strong correlations ($\phi >0.5)$ are, unsurprisingly, (1) planarity and no planarity; and (2) emptiness and thinness.

\section{Approach}
\label{sec:approach}

We now describe the CNN architecture and loss functions that we
use to learn the 
attribute predictors and shape embedding. We cast this as multi-task
training and optimize directly for both.  Specifically, the network is trained
using a loss function over all attributes as well as an embedding loss
that encourages instances of the same shape to have the same
representation. The former lets us model the attributes that are
currently labeled.  The latter forces the network to learn a
representation that can distinguish sculptures, implicitly modeling
aspects of shape not currently labeled.

\begin{figure}[t]
\includegraphics[width=\linewidth]{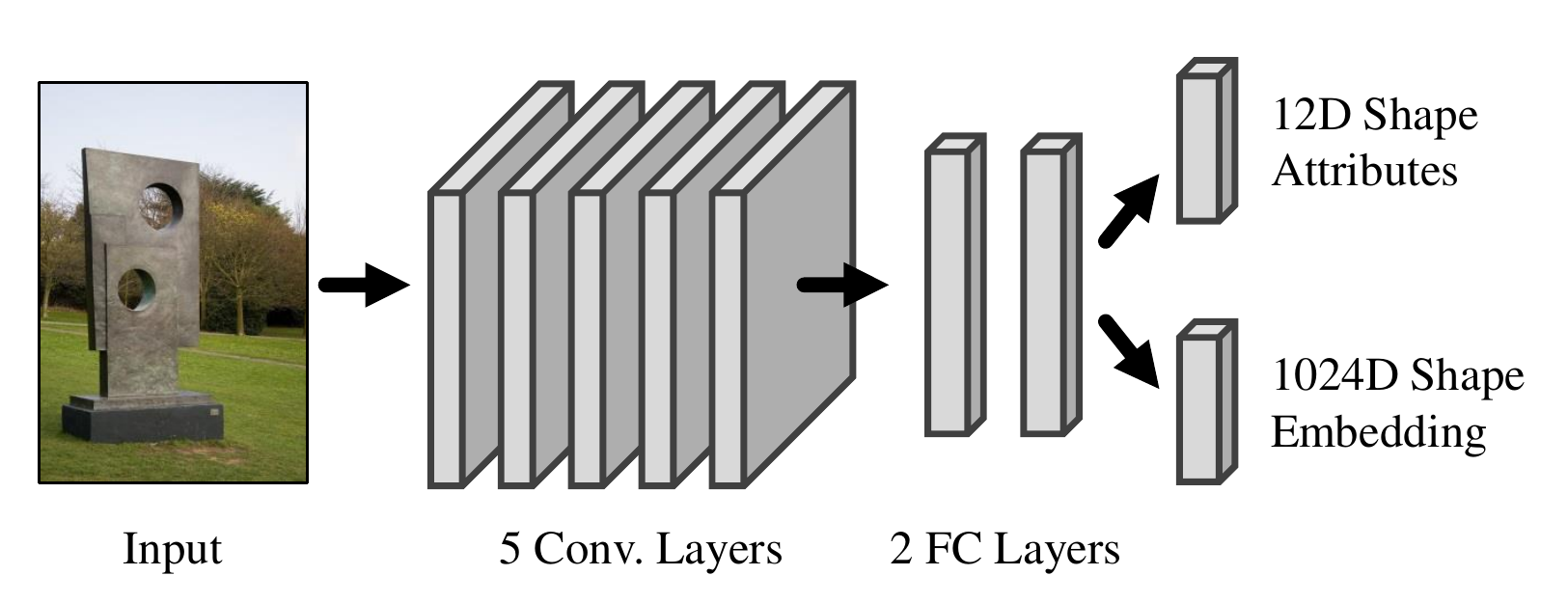}
\caption{The multi-task network architecture, based on VGG-M. After shared layers, the
network branches into layers specialized for attribute classification 
and shape embedding.}
\label{fig:arch}
\end{figure}

\noindent {\bf Network Architecture:} 
    We adapt the VGG-M architecture proposed in \cite{Chatfield14}. We depict the overall
architecture in Fig.\ \ref{fig:arch}: all layers are shared through the last fully connected
layer, fc7. After the 4096D fc7, the model splits into two branches, one for attributes, the other for
embedding. The first is an affine map to $12$D followed by independent sigmoids, producing
$12$ separate probabilities, one per attribute. The second projects fc7 to a $1024$D embedding
which is then normalized to unit norm.
 
We directly optimize the network for both outputs, which allows us to obtain strong 
performance on both tasks. The first loss models all the attributes with a cross-entropy
loss summed over the valid attributes. Suppose there are $N$ samples and 
$L$ attributes, each of which can be $1$ or $0$ as well as $\emptyset$ to
indicate that the attribute is not labeled; the loss is 
\begin{equation}
\small
L(Y,P) = \sum_{i=1}^N \sum_{\substack{l=1\\Y_{i,l} \not =
\emptyset}}^L Y_{i,l} \log(P_{i,l}) + (1-Y_{i,l}) \log(1-P_{i,l}),
\label{eqn:lossAttr}
\end{equation}
for image $i$ and label $l$, 
where we denote the label matrix as $Y_{i,l} \in \{0,1,\emptyset\}^{N,L}$ and
the predicted probabilities as $P_{i,l} \in [0,1]^{N,L}$.  The second loss is an embedding
loss over triplets as in \cite{Wang15b,Schroff15,Schultz04}.
Each triplet $i$ consists of an anchor view of one 
object $x_i^a$, another view of the same object $x_i^p$, as well as a view of a different 
object $x_i^n$. The loss aims to ensure that two
images of the same object are closer in feature space compared to another
object by a margin:
\begin{equation}
\sum_{i=1}
\max \left (
D(x_i^a,x_i^p) - D(x_i^a,x_i^n) + \alpha, 0 \right )
\label{eqn:lossEmb}
\end{equation}
where $D(\cdot,\cdot)$ is squared Euclidean distance.
We generate triplets in a
mini-batch and use soft-margin violaters 
\cite{Schroff15}. 

We see a number of advantages to multi-task learning. It yields a network that can
both name attributes it knows about and model
the 3D shape space implicitly. Additionally, we found it to
improve learning stability, especially compared
to individually modeling each attribute. 

\noindent {\bf Configurations:} We explore two configurations 
to validate that we are really learning about 3D shape. Unless otherwise
specified, we use the system described above, {\it Full}. However,
to probe what is being learned in one experiment, we also learn a network that only optimizes the
attribute Loss (\ref{eqn:lossAttr}), which we refer to as {\it Attribute-Only}.

\noindent {\bf Implementation Details:} 
{\it Optimization:} We use a standard stochastic gradient descent plus momentum approach with
a batch size of 128. We balance the two losses by multiplying the triplet loss
by $3$, which was determined by optimizing each loss independently.
{\it Initialization:} We initialize the network using the model from \cite{Chatfield14} which was 
pre-trained on image classification \cite{ILSVRC15}.
{\it Parameters:} We use a learning rate of $10^{-4}$ for the pre-trained layers, and 
$10^{-3}$ and $10^{-2}$ for classification and embedding layers respectively.
We set the margin $\alpha$ to 0.1; we found that too-large margins led to poor optimization.
{\it Augmentation:} At training time, we use random crops, flips, and color jitter.
At test time, we sum-pool over multiple scales, crops and flips as in
\cite{Chatfield14}.

\begin{figure*}[t]
\includegraphics[width=\linewidth]{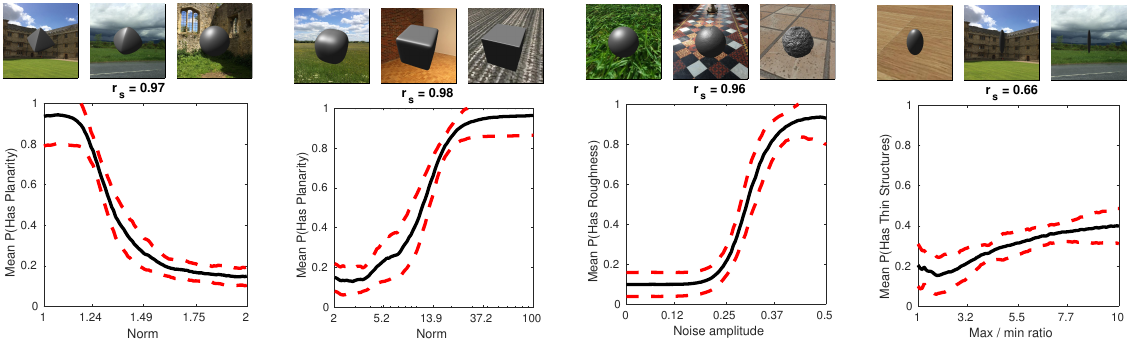} 
\caption{{\bf Plots of predicted attributes vs parameters.}
For each stimulus, we show a plot of the mean predicted shape attribute
against the relevant parameter, sample images at the $10$th, $50$th, and
$90$th percentiles of the parameter, and give the rank correlation. 
Red error bars indicate the standard
deviation after centering the per-background responses at zero.
}
\label{fig:synthetic_results}
\end{figure*}

\section{Analysis by Synthesizing Stimuli}
\label{sec:analysissynthesis}

Interpreting results on pre-captured in-the-wild data can be challenging because
the cause of two images being interpreted differently could be due to any
number of changes between the images. We therefore first analyze our results
in a controlled setting, via synthetic data, where all underlying factors
of an image are tightly controlled. Synthetic data offers an opportunity
to systematically analyze a model's behavior since it enables one to be sure that
only one parameter changes between two images and to control that change. 
This technique was inspired by past work \cite{Zoran15} that probed network
response as a function of 2D patch transformations and simple variations, which in turn
was inspired by human psychophysics. Here, we use a 3D graphics engine
to generate the stimuli, and thus we can modulate properties of the underlying 3D shape
as opposed to 2D transformations.

This approach is complementary to the more commonly used
localization analysis such as \cite{Zhou2016,Simonyan14a} 
(which we perform later in Section \ref{sec:saliency}).
In localization, the goal is to identify parts of a particular image that especially
contribute to a final decision; interpreting these parts or activations is then done post-hoc 
by humans. Here, since all factors of variation can be exactly controlled,
we can examine network response as a function of these factors. The primary disadvantage 
is that it requires a good synthetic model in which the factors of interest can be controlled.
Nontheless, we see three compelling benefits to analysis by synthesizing stimuli: 
(a) it functions as a controlled experiment and can conclusively identify the factor responsible
for a change in the data; (b) it can characterize subtle global changes,
for instance the slight flattening of a sphere; and (c) it enables experiments that
are practically impossible with real images, such as fixing shape and changing texture
or creating hybrid stimuli that combine conflicting cues from two different shapes.

We begin by defining our stimuli, which consist of parameterized deformations
of the unit-norm ball. We then test how well the network learned on sculpture
can interpret these deformations, providing verification that the
network has learned the properties of interest. Finally, having defined our stimuli
and verified that they are being interpreted correctly, we analyze how sensitive
our network is to various cues, using planarity as our property of interest
due to the large literature on curvature perception (e.g., \cite{Biederman87,Johnston94,Koenderink92}).
Our results show that the network is simultaneously incorporating a variety of signals
ranging from mathematically-modelable shape cues such as shading and contours 
to data-driven correlations between shape, color, and texture. 

\subsection{Stimuli}

We use three synthetic stimuli consisting of 
the deformation of a unit-norm ball; each stimuli is parameterized
by a single parameter $p$:
(i) {\it $L_p$:} the $L_p$ ball, $\{\xB \in \mathbb{R}^3: ||\xB||_p = 1\}$, 
i.e., $p = 1$ is a octahedron, $p = 2$ is a sphere, and $p = \infty$ is
a cube. We split this stimulus into two stimuli to ensure
a monotonic relationship between planarity and the parameter:
a stimulus with $p \in [1,2]$, linearly spaced, and 
one with $p \in [2,100]$, logarithmically spaced.
(ii) {\it Noise:} the unit sphere with the radius at each vertex
additively displaced by a fixed noise pattern generated with fractal Brownian
noise. We vary the magnitude of this noise $p \in [0,0.5]$. 
(iii) {\it Oval:} A sphere with the $X$ and $Z$ axes
scaled by a factor $p \in [1,\frac{1}{10}]$. 
Each tests a different attribute: $L_p$ tests questions of planarity; {\it Noise} 
test questions of roughness; and {\it Oval} tests cubic aspect ratio and 
thinness.

Each geometry was generated with $\approx 90K$ vertices and
then rendered with a gray specular material under a soft ambient
light and a single directional light source
using the code of \cite{Aubry14}. Finally, each rendering
was composited on top of 10 images of open spaces depicting indoor
and outdoor spaces and no salient objects. We use multiple backgrounds
to preclude effects due to any one particular background (e.g., inadvertent camouflaging). 

\subsection{Accuracy Experiments}

After training the network described in Section \ref{sec:approach}, 
we first verify that the network can interpret these stimuli correctly.
In addition to providing additional confirmation that the network
has learned the actual properties, any subsequent analysis is useless
unless the network interprets the stimuli correctly.
We should note that this is not guaranteed: as \cite{Su15} points out, 
there is a considerable domain shift between real and synthetic images.

\noindent {\bf Quantitative criteria:} 
Our stimuli all satisfy the property that increasing a parameter $p$
increases the presence of an attribute in the shape.
We can thus quantify performance by evaluating the correlation between
attribute predictions and parameter $p$. Since the relationship
is not necessarily linear, we use Spearman's rank correlation 
$r_s$, which characterizes whether there is a monotonic relationship: $1$ indicates
perfect rank correlation; $0$ indicates no correlation. Since the
background alters the predicted attribute, we analyze
per-background and report the average across backgrounds.

\noindent {\bf Results:} We show a plot of the predicted shape attributes
against the varied parameter $p$ in Fig.~\ref{fig:synthetic_results}, as well
as the rank correlation. The black line indicates the average across backgrounds.
Directly computing the standard deviation across backgrounds mixes the actual
uncertainty with a per-background bias that each background may introduce: 
the backgrounds with planar textures are viewed as more planar by the network, for instance. 
We thus compute an updated standard deviation after centering each background's graph at zero
and report $1$ standard deviation with a red error bar.

Despite the great dissimilarity between these stimuli and the data on which the
model was trained, the network successfully generalizes well and a clear trend
emerges in each case. If we repeat the analysis on all the backgrounds pooled together, this trend remains similar,
and the rank correlations in each case decrease by an average of only $0.1$.

\begin{figure}
\includegraphics[width=\linewidth]{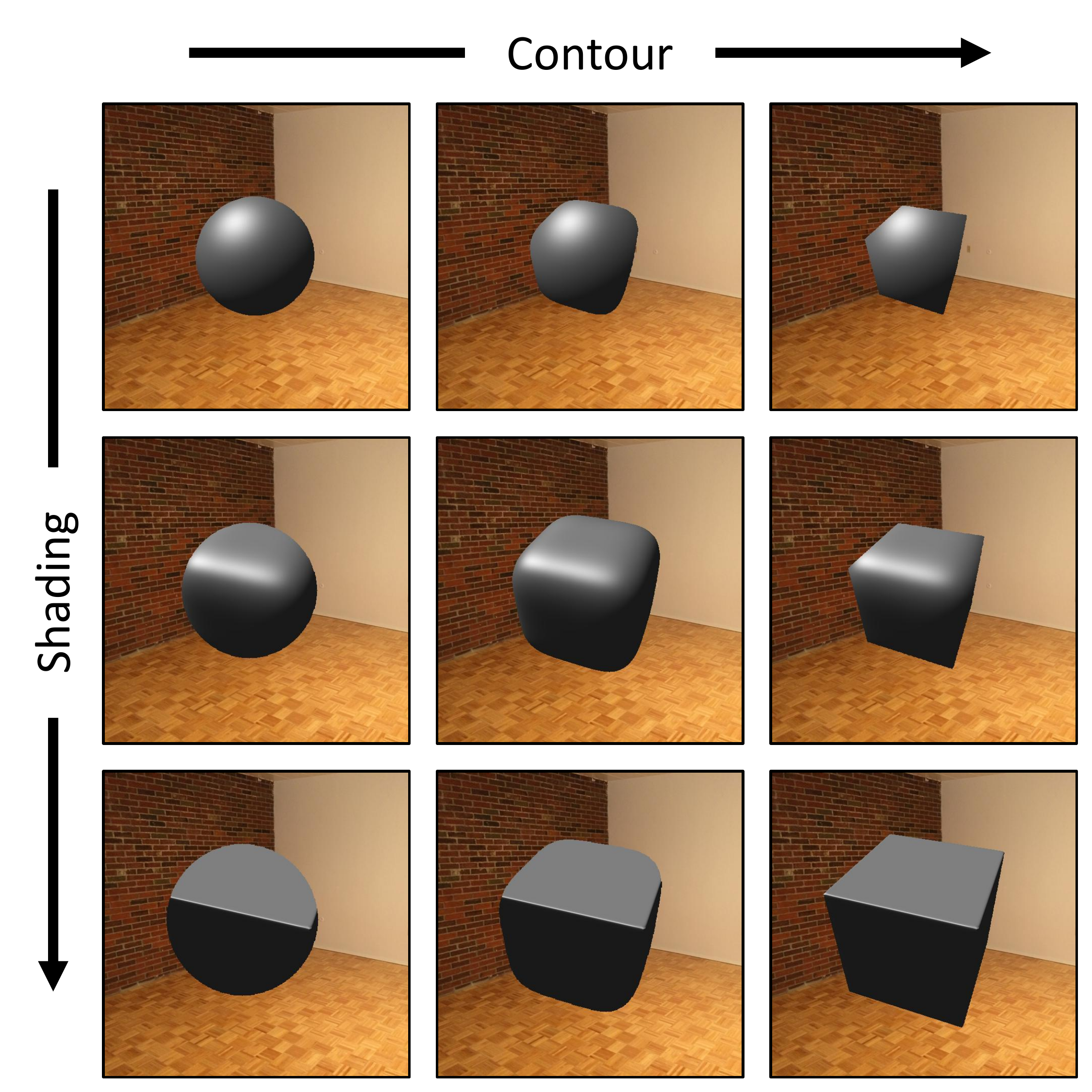}
\caption{Examining the role of contour and shading cues in the output
of the network. The contour and shading changes from sphere to cube going 
along the horizontal and vertical directions respectively. The network
perceives stimuli closer to the bottom right as more planar.}
\label{fig:conflictingcues}
\end{figure}

\subsection{Results with Conflicting Cues}

There are a variety of cues by which people and machines can see 3D, so an
important question is how the network is doing it.  For instance:
the results on PASCAL showed that a network trained on sculptures
could accurately identify non-planar trains. It is not clear, however, what cues
were used. Now that we have showed that the network interprets the stimuli correctly,
we aim to address this question.

The synthetic stimuli let us analyze this question via composite objects that
have conflicting cues, similar to cue combination techniques used in human subjects
\cite{Bulthoff89,Norman95,Madison01,Welchman05}. Here, we use this to analyze the role of contour and shading
cues in predicting the planarity of an object by creating objects that
combine the contours and shading cues of a sphere and cube  -- for instance, 
a sphere that has the occluding contour of a cube. By shading, we mean the 
change in intensity caused by the projection of the light onto a particular
shape. We show some examples of
these objects in Fig.~\ref{fig:conflictingcues}: each row or column depicts a fixed
shading (row) or occluding contour (column); the original stimulus goes along
the diagonal. In the original stimulus, the cues were varied jointly, but
we can also fix one cue and vary the other.

\begin{table}
\centering
\caption{How {\it has planar surfaces} changes as contour and shading 
vary or are held constant as a sphere or cube.
Changing both cues simultaneously 
naturally causes the strongest response; changing each cue
while holding the other constant causes a change; 
and varying shading produces a stronger response of the two cues. }
\label{tab:conflictingcues}

\begin{tabular}{cccccc} \toprule
\bf Shading & Varying & \multicolumn{2}{c}{Varying} & Sphere & Cube \\
\bf Contour & Varying & Sphere & Cube               & \multicolumn{2}{c}{Varying} \\ \midrule
$r_s$       & 0.98 & 0.96 & 0.97 & 0.50 & 0.36 \\
Range       & 0.85 & 0.71 & 0.73 & 0.15 & 0.14 \\
\bottomrule
\end{tabular} 
\end{table}

\noindent {\bf Results:} Both cues are being used by the network, but shading cues appear to dominate
contour ones. We quantify this by correlation and range of responses: if
a cue is being used, the perceived planarity should be correlated with the
change in cue (i.e., more planar contours produce more planar perceptions); 
the range of values taken when a cue is varied indicates how 
heavily the cue is used. We show both metrics in Table \ref{tab:conflictingcues} 
in the case where we vary one or more cues. It can be seen that: both cues are used; 
the strongest response occurs when both are varied at the same time; and if
only shading produces a far stronger response than contour.
We also found that the contour cue was inconsistent in its effectiveness across
backgrounds, presumably due to varying difficulty in finding the precise
contour.

\subsection{Sensitivity to Light}

If shading cues are important for interpreting shape, 
then it may be sensitive to the lighting conditions. Here, we experimentally examine
this with a set of 100 lighting setups. These have
randomized count ($1$--$6$), locations, and colored intensities; results
are similar with grayscale lighting, but lower variance in general.

\noindent {\bf Results:} One way to evaluate how sensitive the network is to lighting is to quantify how
lighting and predicted planarity vary together. For instance, ideally 
we ought to see the same strong correlation between underlying shape and predicted shape,
even under extreme and unrealistic lighting. We find this to be true: for fixed background
and lighting, there is still an average $0.9$ rank correlation between the predicted
and actual shape; occasional failures happen with harsh overexposing lighting that obscures details.  
Similarly, if we fix the input stimulus, we ought to see little variance as we change the lighting.
Unambiguous stimuli (octahedra/spheres/cubes) were interpreted consistently: the 
standard deviation across the lightings ranges from $0.06$ to $0.08$ on these
stimuli (as reference the difference between the average cube and sphere interpretation 
is $0.7$). Ambiguous stimuli had a higher variance, but all had standard
deviations below $0.21$.

An important case are catastrophic errors where lighting
radically changes shape interpretation.
We quantify this by examining how frequently octahedra and
cubes were predicted as less planar than spheres. This never occurred when
lighting was set identically for the stimuli. Changing lighting independently for 
the two stimuli cause mistakes in a handful ($37$) of cases out
of the $200K$ possible pairs ($<0.02\%$). Most were caused by lighting
hiding one stimulus and revealing the other (e.g., 
green light on one stimulus and deep purple on another, both in front
of green hills).

\begin{figure}
\includegraphics[width=\linewidth]{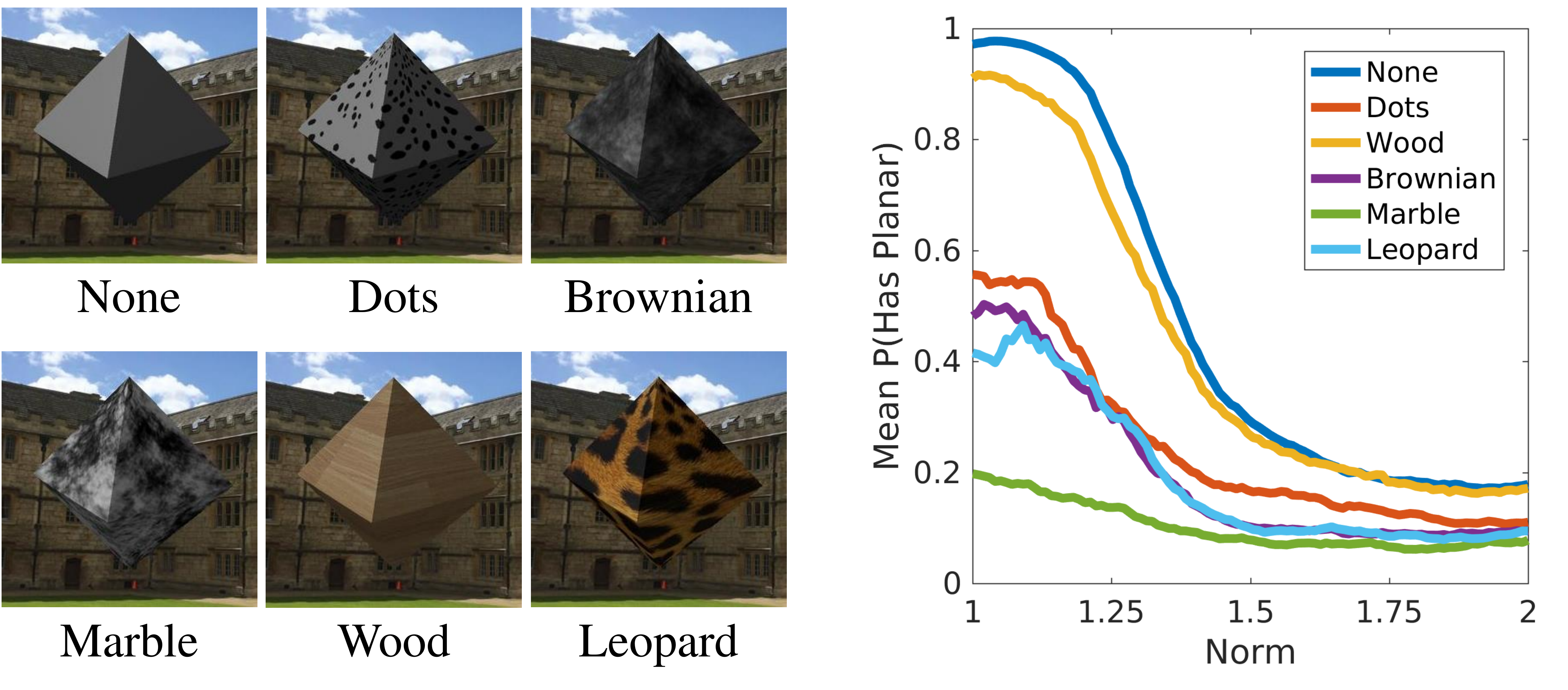}
\caption{Texture variations and their effect on the perception of the $L_p$ stimulus.
Marble-like texture changes the interpretation, although
changing shape also changes the prediction irrespective of texture.}
\label{fig:texture}
\end{figure}

\subsection{Sensitivity to Texture}

We next examine whether the
network has learned any correlations between texture and shape. 
We try six texture options: (1) {\it None}, the original material; (2) {\it Dots},
randomized ellipses; (3) {\it Brownian}, fractal Brownian noise; (4) {\it Marble}, which is the Brownian stimulus 
histogram equalized, which yields a marble-like effect; (5) {\it Wood}; and (6) {\it Leopard print}. 
Dots tests a simple synthetic stimuli; Brownian and Marble are useful since
they are simple transformations of each other; Wood and Leopard test
textures that are not simple mathematical processes.

\noindent {\bf Results:} We show the six stimuli and the response curve, averaged over backgrounds, for
each texture on the $L_p$, $p \in [1,2]$ stimulus in Figure \ref{fig:texture}.
First, we note that, just as in the case of lighting variations, the network still
produces the correct response to the stimulus for any particular texture: the
average rank correlation between perceived geometry and underlying geometry is $0.96$.
Second, the differences between the curves suggest that the network has learned
to exploit correlations between texture and shape: for instance, simply
histogram-equalizing the Brownian stimulus to make it look like marble changes
the overall likelihood of planarity. This is presumably due to the fact that most marble 
statues are non-planar. Nonetheless, although texture modulates the planarity response,
the actual shape continues to control it. This suggests that the network 
factors in natural correlations between texture and shape, but is not completely
controlled by it.

\subsection{Conclusions}

In this section, we have made steps towards characterizing the behavior of a
CNN trained to predict shape attributes via synthetic stimuli. In the process, we have
provided evidence that the network is sensitive to the correct factors of variation
and relatively insensitive to spurious signals such as lighting. Moreover, we
have demonstrated that it uses classic cues to shape such as contours and
shading but that it relies more on shading cues. Finally, we have demonstrated
that the network has learned to exploit correlations between materials and shape

\begin{table*}

\caption{Area under the ROC curve. Higher is better. Our approach achieves strong
performance and outperforms the baselines by a large margin.}
\label{tab:predictionAll}
\centering
\normalsize
\newcommand{\icwidth}{0.30in}
\newcommand{\icdist}{~~~}
\newcommand{\igdist}{~~~}
\definecolor{LightGreen}{rgb}{0.75,1,0.75}
\definecolor{LightRed}{rgb}{1,0.75,0.75}
\definecolor{LightBlue}{rgb}{0.75,0.75,1}
\definecolor{Gray}{rgb}{0.75,0.75,0.75}

\noindent
\begin{tabular}{@{}l@{~~~~}
                >{\columncolor{LightBlue}}c@{\icdist}>{\columncolor{LightBlue}}c@{\icdist}>{\columncolor{LightBlue}}c@{\icdist}>{\columncolor{LightBlue}}c@{\igdist}c@{\igdist}
                >{\columncolor{LightGreen}}c@{\icdist}>{\columncolor{LightGreen}}c@{\igdist}c@{\igdist}
                >{\columncolor{LightRed}}c@{\icdist}>{\columncolor{LightRed}}c@{\icdist}>{\columncolor{LightRed}}c@{\icdist}>{\columncolor{LightRed}}c@{\icdist}
                >{\columncolor{LightRed}}c@{\icdist}>{\columncolor{LightRed}}c@{\icdist~~}c>{\columncolor{Gray}}c} \toprule
                
                & \multicolumn{4}{c}{\bf Curvature~~~~} & ~~
                & \multicolumn{2}{c}{\bf Contact~~~~} & ~~
                & \multicolumn{6}{c}{\bf Occupancy~~~~~} & \multicolumn{2}{c}{~}
                \\ 
                & \includegraphics[width=\icwidth]{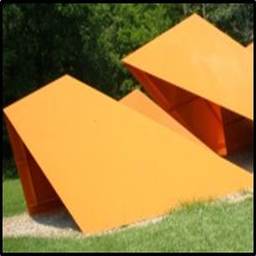} 
                & \includegraphics[width=\icwidth]{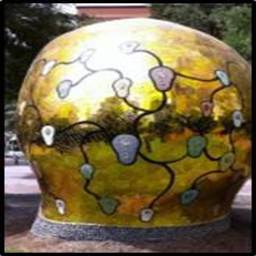} 
                & \includegraphics[width=\icwidth]{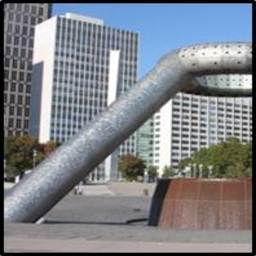} 
                & \includegraphics[width=\icwidth]{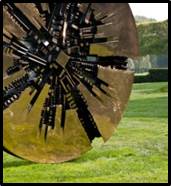} &
                & \includegraphics[width=\icwidth]{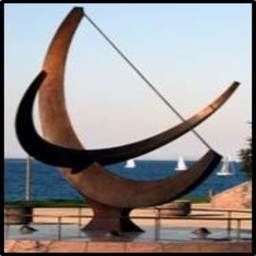} 
                & \includegraphics[width=\icwidth]{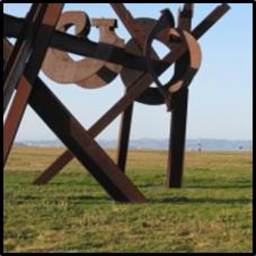} &
                & \includegraphics[width=\icwidth]{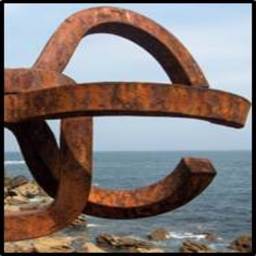} 
                & \includegraphics[width=\icwidth]{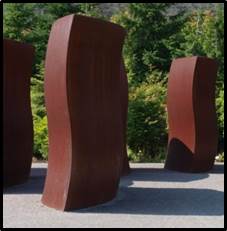} 
                & \includegraphics[width=\icwidth]{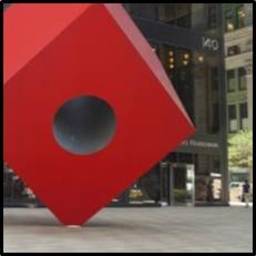} 
                & \includegraphics[width=\icwidth]{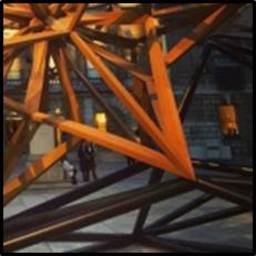} 
                & \includegraphics[width=\icwidth]{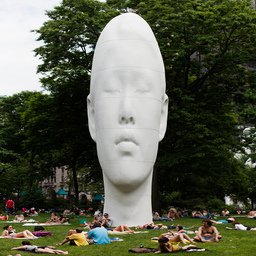} 
                & \includegraphics[width=\icwidth]{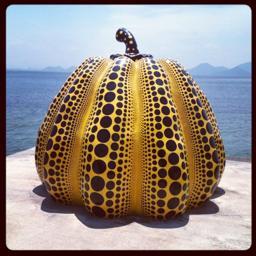} 
                & \multicolumn{2}{c}{~}
                \\

Method ~~~~~~~~~
                & Plan & $\neg$\,Plan & Cyl & Rough &
                & P/L & Mult &
                & Emp & Mult & Hole  & Thin & Sym & Cubic &
                & Mean 

\\ 
\cite{Barron15}\,+\,\cite{Bo11}~~~~&
64.1 & 63.4 & 51.2 & 61.3 & & 61.1 & 61.6 & & 66.5 & 52.8 & 56.0 & 63.5 & 56.2 & 55.7 & & 59.4 \\
\cite{Eigen14}\,+\,\cite{Bo11} & 
64.6 & 61.0 & 50.6 & 60.6 & & 57.5 & 60.9 & & 65.2 & 55.7 & 52.4 & 65.7 & 57.2 & 51.2 & & 58.5 \\
\cite{Barron15}\,+\,\cite{Gupta14} &
70.0 & 64.4 & 53.1 & 63.9 & & 63.6 & 64.8 & & 73.7 & 56.4 & 54.1 & 69.7 & 60.2 & 56.2 & & 62.5 \\
\cite{Eigen14}\,+\,\cite{Gupta14} & 
67.5 & 61.8 & 51.9 & 64.8 & & 58.5 & 64.8 & & 71.5 & 57.8 & 52.4 & 67.7 & 59.4 & 56.1 & & 61.2 \\
Proposed & 
\bf 82.8 & \bf 77.2 & \bf 56.9 & \bf 76.0 & & \bf 74.4 & \bf 76.4 & & \bf 87.0 & \bf 60.4 & \bf 69.3 & \bf 85.8 & \bf 60.8 & \bf 60.3 & & \bf 72.3 \\

\bottomrule
\end{tabular}
\end{table*}

\section{Experiments}
\label{sec:experiments}

\begin{figure}[t]
\includegraphics[width=\linewidth]{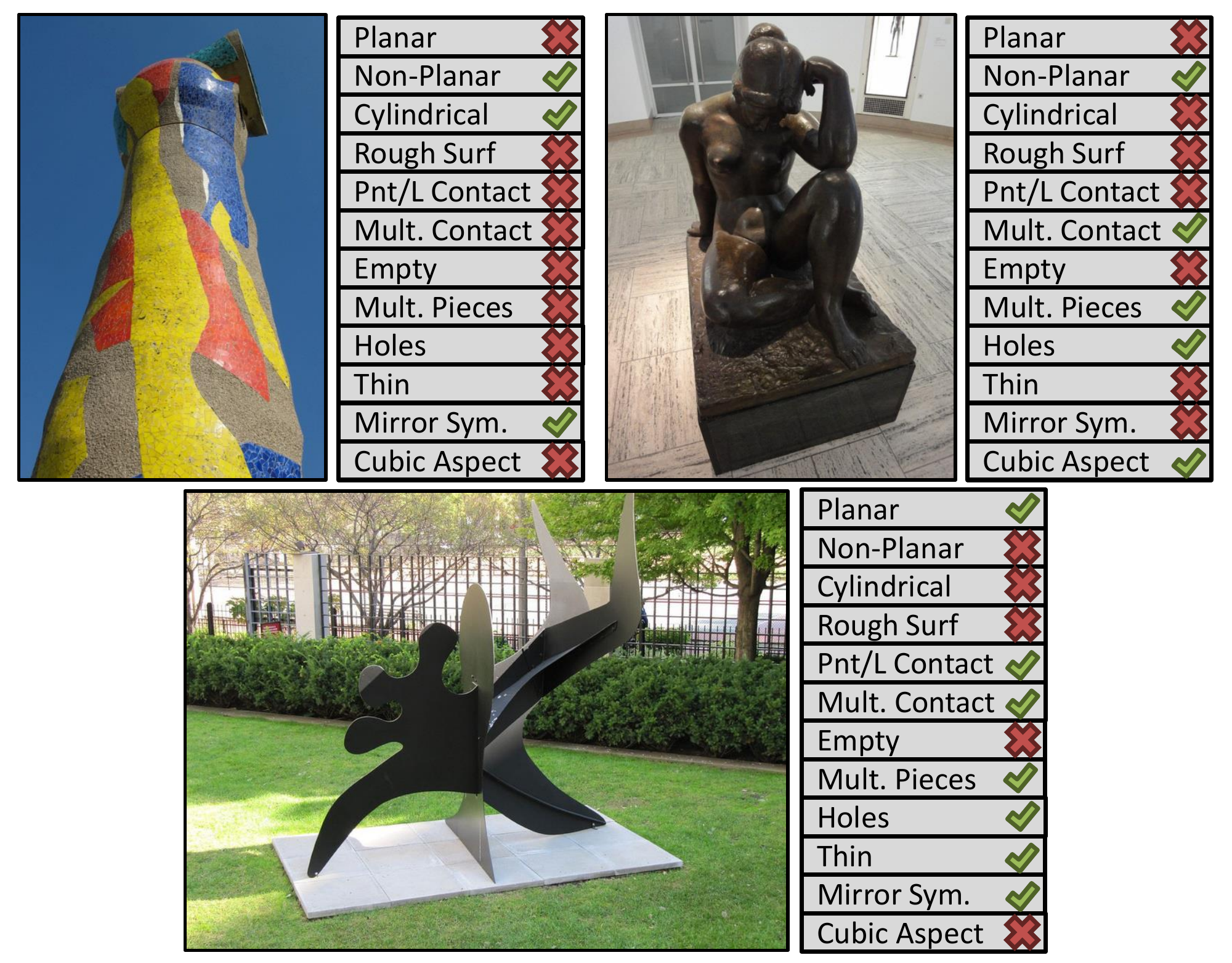}
\caption{\small Thresholded predictions for all attributes on test images. The system
has never seen these sculptures or ones by the artists who made them, but generalizes
successfully.}
\label{fig:QualitativeAttribute}
\end{figure}

Having analyzed the behavior of the network on synthetic data, we now evaluate it
on our real data. We describe a set of experiments to investigate both the performance
of the learnt 3D shape attribute classifiers, {\em and} what has been learnt.
We aim to answer two basic questions in this section:
(1) how well can we predict 3D shape attributes from a single image? 
and (2) are we actually predicting 3D properties or a proxy property that 
correlates with attributes in an image?
To address (1) we evaluate the performance on the Sculpture Images Test set,
and also compare to alternative approaches that 
first predict a metric 3D representation
and then derive 3D attributes from that  (Sec.~\ref{sec:exp_pred}).
We probe (2) in a variety of ways. First, we examine the regions of the image
responsible for the predictions in Sec~\ref{sec:saliency}. Second, we evaluate 
the learnt representation on a different task -- determining 
if two images from different viewpoints are of the same object or not
(Sec.~\ref{sec:exp_rot}). Third, we evaluate how well the 3D 
shape attributes trained on the Sculpture images generalize to non-sculpture data, in particular to predicting
shape attributes on PASCAL VOC categories (Sec.~\ref{sec:exp_pascal}). 
Finally, we probe the model with a set of synthetic stimuli in Section \ref{sec:analysissynthesis}.

\subsection{Attribute Prediction}
\label{sec:exp_pred}

We first evaluate how well 3D shape attributes can be estimated from
images. Here, we report results for our full network. Since our dataset is
large enough, the attribute-only network does similarly.
We compare the approach proposed in this paper (which directly
infers holistic attributes) to a number of baselines that are depth orientated,
and start by computing a metric depth at every pixel.

\noindent {\bf Baselines:} The baselines start by 
estimating a metric 3D map, and then attributes are extracted from this 
map.  We use two recent methods for estimating depth from single images with code available:
a CNN-based depth estimation technique \cite{Eigen14} and an intrinsic images
technique \cite{Barron15}. Since \cite{Barron15} expects a mask, we 
use the segmentation used for collecting the dataset (in Sec.~\ref{sec:clusters}).
One question is: how do we convert these depthmaps into our attributes?
Hand-designing a method is likely to produce poor results. We take a data-driven approach
and treat it as a classification problem. We use two approaches
that have produced strong performance in the past. The first is a linear SVM on 
top of kernel depth descriptors \cite{Bo11}, which convert the depthmap into
a high-dimensional vector incorporating depth configurations
and image location. The second is the HHA scheme \cite{Gupta14}, which converts
the depthmap into a representation amenable for fine-tuning a CNN; in this
case, we learn the attribute CNN described in Section~\ref{sec:approach}.

\noindent {\bf Evaluation Criteria:} 
Each method produces a prediction scoring how much the image has
the attribute. We characterize the predictive ability of these scores
with a receiver operator characteristic (ROC) over the Sculpture
images test set.  This enables comparison {\it across attributes}
since the ROC is unaffected by class frequency
\cite{Fawcett06}. We summarize scores with
the area under the ROC curve (AUROC).

\begin{figure}[t]
\includegraphics[width=\linewidth]{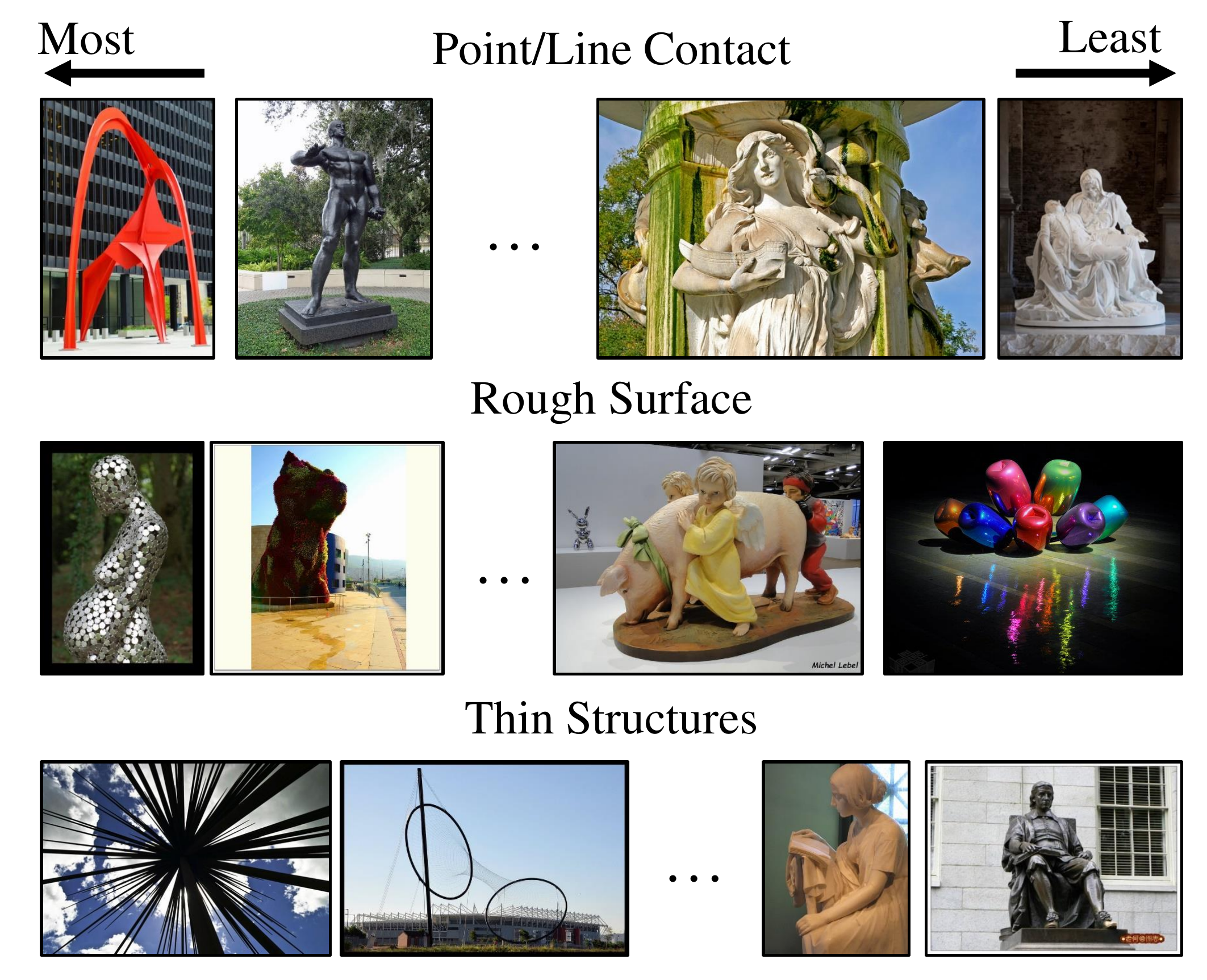}
\caption{\small Test images sampled at the top, 95$^{\textrm{th}}$, 
5$^{\textrm{th}}$ percentiles and lowest percentile with respect to three attributes.}

\label{fig:qualitativePred}
\end{figure}

\noindent {\bf Results:} 
Fig.~\ref{fig:QualitativeAttribute}
shows thresholded predictions of all of the attributes on a few sculptures. To help visualize
what has been learned,  we show automatically sampled results
in Fig.~\ref{fig:qualitativePred}, sorted by the predicted presence of attributes. 

We report quantitative results in Table \ref{tab:predictionAll}. On an absolute
basis, certain attributes, such as planarity and emptiness, are easier than
others to predict, as seen by their average performance; harder ones
include ones based on symmetry and aspect ratio, which may require a global
comparison across the image, as opposed to aggregation of local judgments.

In relative terms, our approach out-performs the baselines, with especially large
gains on planarity, emptiness, and thinness. Note that
reconstructing thin structures is challenging even with multi-view stereo as input
and typically requires specialized handling \cite{Ummenhofer2013};
an approach based on depth-prediction is thus likely to fail at reconstruction, and
thus on attribute prediction. Instead, our system directly recognizes that the object is thin 
(e.g., Fig.~\ref{fig:QualitativeAttribute} bottom). 
Fig.~\ref{fig:qualitativePred} shows that frequently, the instances that least have an 
attribute are the negation of the attribute: for
example, even though many other sculptures are not rough, the least
rough objects are especially smooth.

The system's mistakes primarily occur on images where it is uncertain: 
sorting the images by attribute prediction and re-evaluating on the top and bottom 25\% of the images
yields a substantial increase to 77.9\% mean AUROC; using the top and bottom $10\%$ yields an increase to 82.6\%.

Throughout, we fix our base representation to VGG-M \cite{Chatfield14}. 
Switching to VGG-16 \cite{Simonyan14c} gives an additional boost: the mean
increases from 72.3 to 74.4 and $1/3$ of the attributes are predicted with AUROCS 
of $80\%$ or more.

\subsection{Saliency Maps}
\label{sec:saliency}

\begin{figure*}
\includegraphics[width=\linewidth]{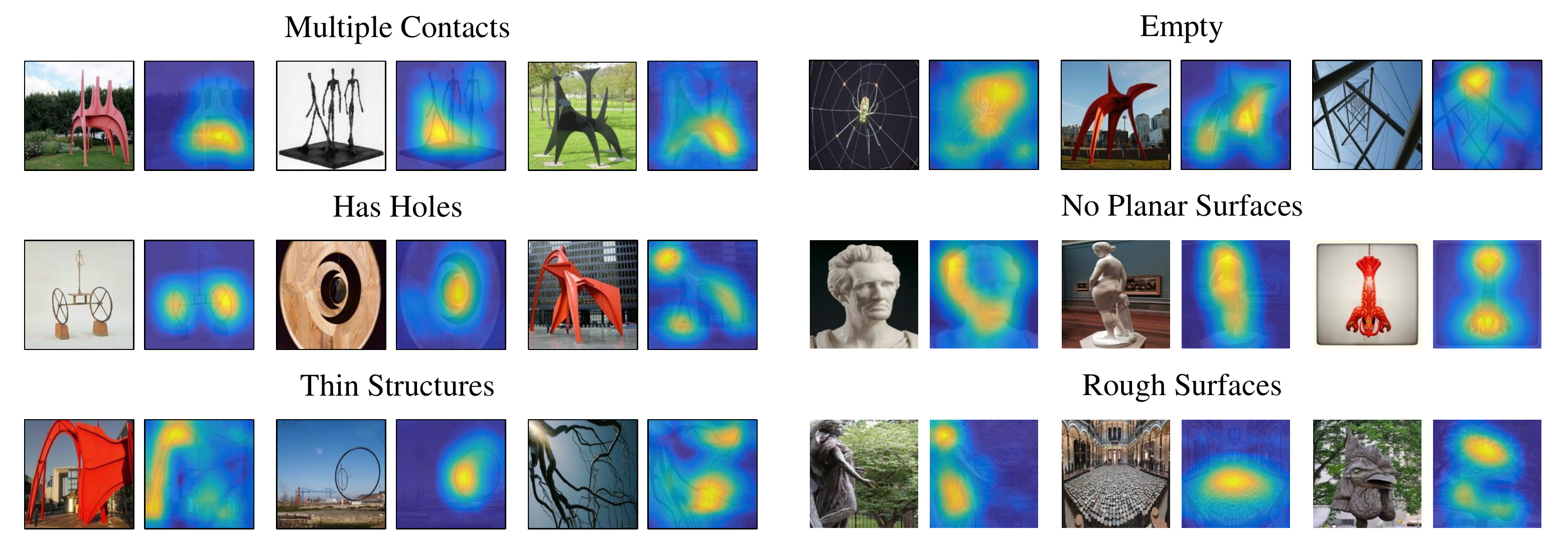}
\caption{Class Activation Maps for six of the shape attributes. For each attribute, we select 
an image that the network predicts has high presence of an attribute (in top 1K scoring images)
and overlay the class activation map for that attribute.}
\label{fig:cam}
\end{figure*}

As a way of examining the network, we use the class-activation
mapping (CAM) technique from \cite{Zhou2016}.

\noindent {\bf Experimental Setup:} 
Using the CAM technique involves connecting the last convolutional layer 
to the classification weights by average pooling, thus producing a final feature
that has as many channels as the convolutional layer but $1 \times 1$ spatial resolution.
We thus remove both fully connected layers of our VGG-M network and attach an
average pooling layer to the conv5 layer.
Having done this, we retrain the network following identical settings. 

It should be noted that this produces a different network. However, we found
it makes similar decisions to the network trained in Section \ref{sec:exp_pred}:
the average correlation between the retrained and previous networks' activations on
the test set is high ($0.87$); the mean AUROC is slightly 
($0.64\%$) lower (consistent with results reported for other architectures
in \cite{Zhou2016}), and the maximum deviation of any attribute's ROC is $3.9\%$.

\noindent {\bf Results:} We examine saliency by looking at images in the images
which cause the top $1$K strongest predictions for each attribute in the test
set. Fig.~\ref{fig:cam} shows a selection of these for six attributes. The
maps suggest that the network is using the right parts of the image to make its
decision, even in the case of confident mistakes ({\it Has Holes}, right, which
appears to be a hole due to an accidental viewpoint). In the case of
analyzing contact, the network appears to be using the place at which ``legs''
of the sculpture split apart, and employs a similar strategy for the empty
property. Roughness seems driven by rough surfaces, or, judging by the map on the
set of stools, particular texture frequencies. We found that more global
properties, such as mirror symmetry produce results that are more difficult to
interpret.

\noindent {\bf Localization Results:}
We found that the CAM maps localized the sculpture well, suggesting
that irrespective of which part is being used, the sculpture itself is
driving predictions. To quantify this, we examined segmentation on a
set of $40$ hand-segmented images. We treat the CAM maps
as per-pixel predictions of whether the sculpture is in that
pixel and evaluate the predictions by computing an AUROC on a per-pixel basis.
Each CAM map produces at least a 71\% AUROC; when normalized and averaged, 
they together achieve 85\% AUROC.

\subsection{Mental Rotation}
\label{sec:exp_rot}

\begin{figure}
\centering
\includegraphics[width=\linewidth,trim=0in 0.1in 0in 0in]{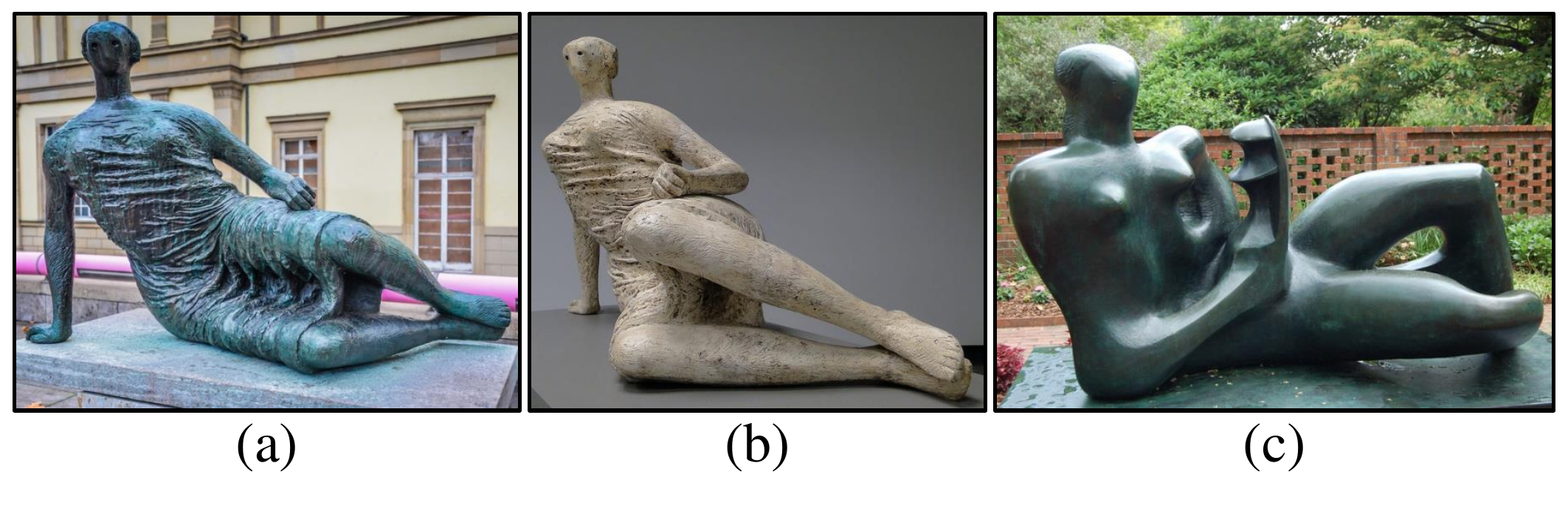}
\caption{In mental rotation, the goal is to verify that (a) and (b) correspond and
(a) and (c) do not. Roughness is a useful cue here. }
\label{fig:rotationExample}
\end{figure}

If we have learned about 3D shape, our learnt representation
ought to encode or {\em embed} 3D shape. But how do we characterize
this embedding systematically? To answer this, we turn to the task of mental
rotation~\cite{Shepard71,Tarr1998} which is the following: 
given two images, can we tell if
they are different views of the same object or instead 
views of different objects? This is a classification task
on the two presented images: for instance, in 
Fig.~\ref{fig:rotationExample}, the task is to tell that (a) and (b) 
correspond, and that (a) and (c) do not.

Note, the design of the dataset has tried to ensure that sculpture
shape is not correlated with location by ensuring that images of a 
particular work come from different locations (since multiple instances of
a work are produced) and 
different materials (e.g., bronze and stone in Fig.~\ref{fig:rotationExample}).

We report four representations: (i) the 1024D
embedding produced by our full network; (ii) the 4096D fc7 layer of the
full network; (iii) the 4096D fc7 layer of the attribute-only network;
(iv) the attribute probabilities themselves from the full network.
If our attribute network is using actual 3D properties, then the attribute network's activations ought 
to work well for the mental rotation task even though it was never trained for it
explicitly. Additionally, the attributes themselves ought to perform well.

\noindent {\bf Baselines:} We compare our approach to (i) the pretrained FC7 from the initialization
of the network and to (ii) IFV \cite{Perronnin10} over the BOB descriptor \cite{Arandjelovic11} that was used
to create the dataset and dense SIFT \cite{Lowe2004}.
The pre-trained FC7 characterizes what has been learned; the IFV representations help characterize 
the effectiveness of the attribute predictions on their own.
We use the cosine distance throughout.

\noindent {\bf Evaluation Criteria:} We adopt the evaluation protocol of 
\cite{LFWTech} which has gained wide acceptance in face verification:
given two images, we use their distance as a prediction
of whether they are images of the same object or not. Performance
is measured by AUROC, evaluated over 100 million of the pairs, of
which $0.9\%$ are positives. Unlike \cite{LFWTech}, positives in the
same viewpoint cluster are ignored: these are too easy 
decisions. 

We further hone in on difficult examples by automatically finding and removing 
easy positives which can be identified with a bare minimum image representation. Specifically,
we remove positive pairs with below-median distance in a
512-vocabulary bag-of-words over SIFT representation. This yields a 
more challenging dataset with $0.3\%$ positives. As mentioned in 
Sec.~\ref{sec:dataset} artists often produce work of
a similar style, and the most challenging examples
are often pairs of images from the same artist (which may or may not be
of the same work).
We call the standard setting {\it Easy} and the filtered setting
with only hard positives {\it Hard}.

\begin{figure}
\centering
\begin{tabular}{cc}
\includegraphics[height=1.5in]{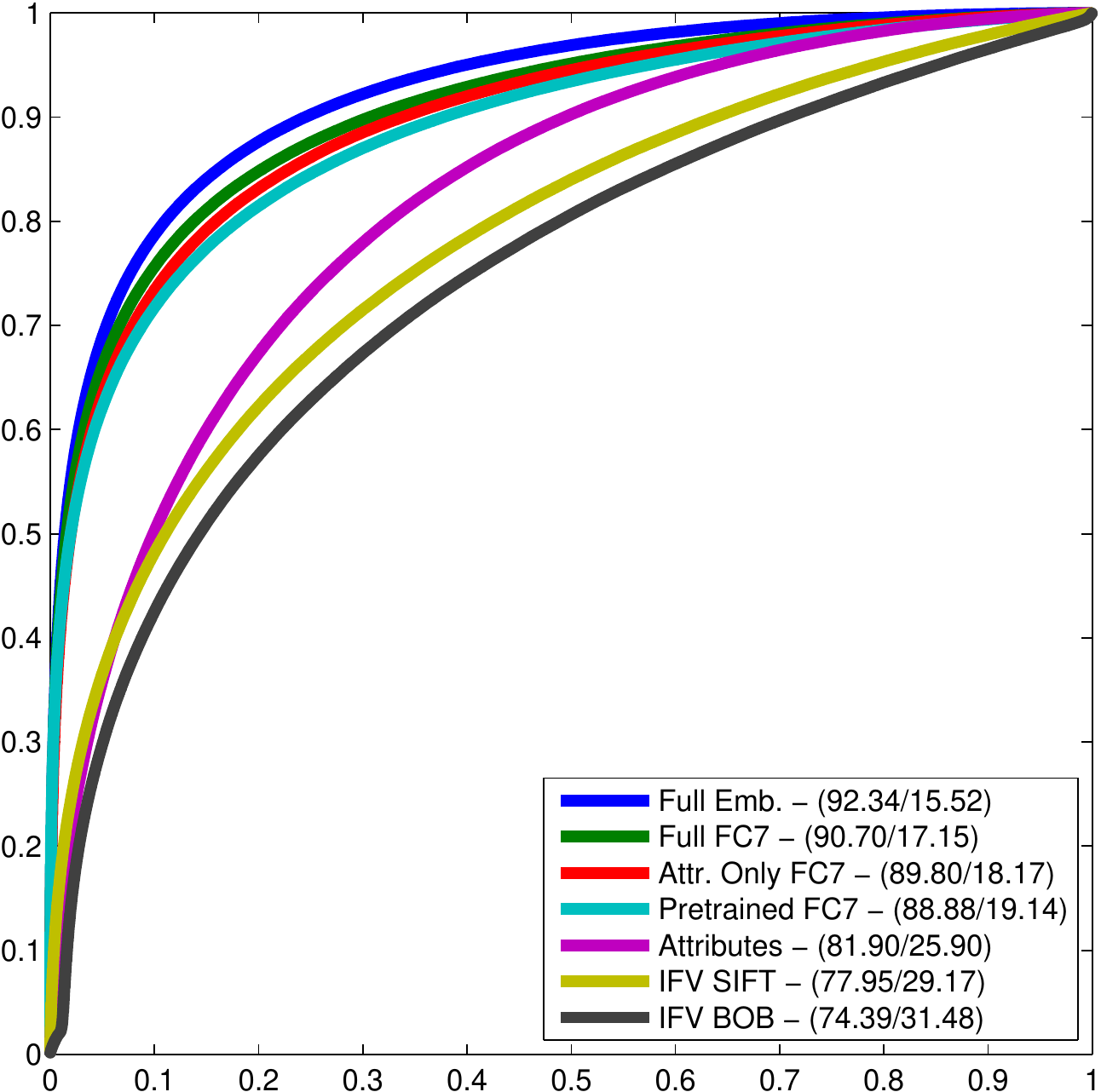} &
\includegraphics[height=1.5in]{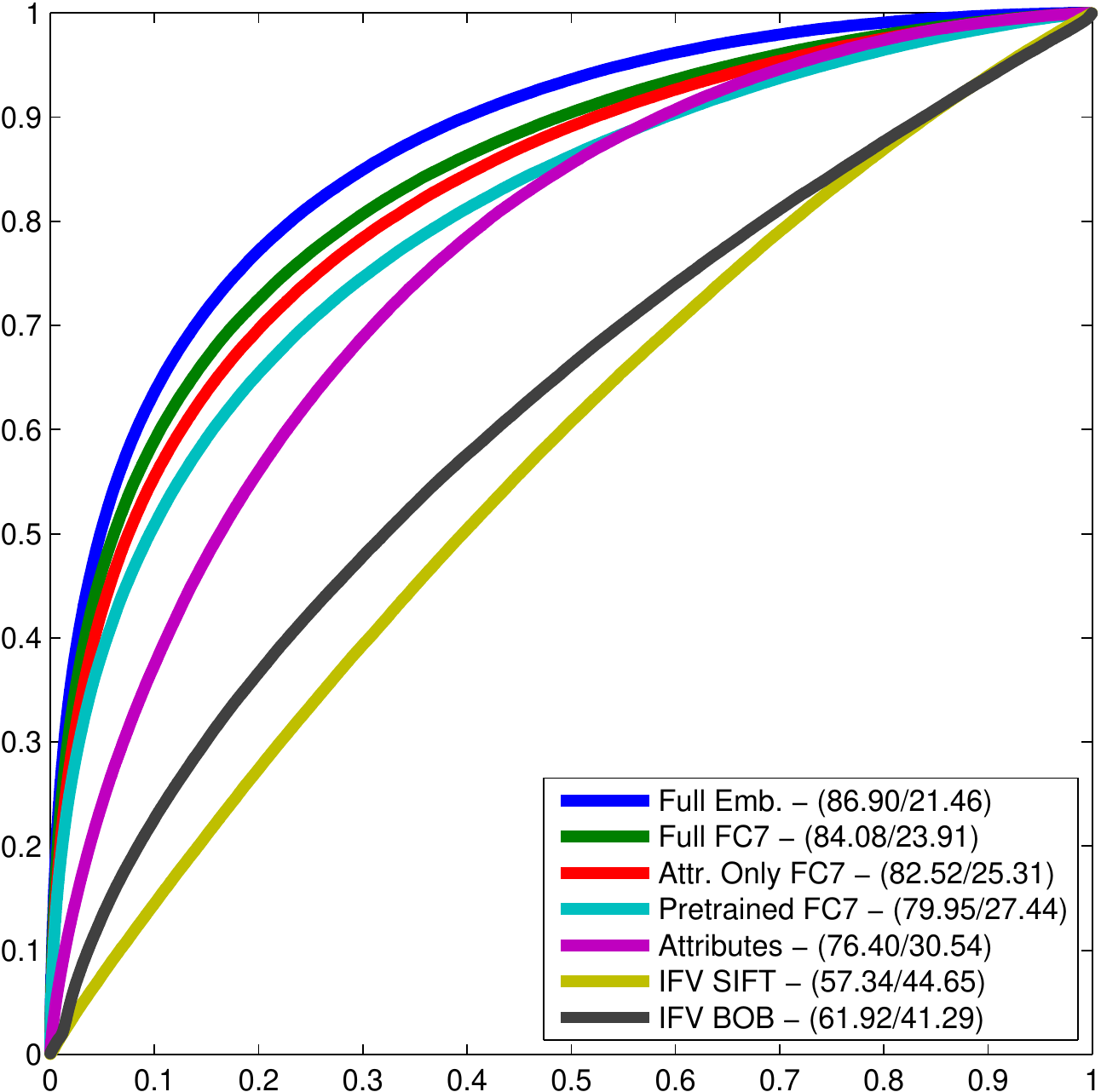} \\
\small (a) Easy Setting & \small (b) Hard Setting \\
\end{tabular}
\caption{Mental rotation ROCs for easy and hard settings. In the legend, we report the AUROC and EER for each method.} 
\label{fig:rotation}
\end{figure}

\noindent {\bf Quantitative Results:} Table~\ref{tab:rotation} and
Fig.\ \ref{fig:rotation} show results for both settings. By themselves, 
the 12D attributes produce strong performance, 3-4\% better than 
IFV representations. The attribute-only network improves over pretraining (by 0.9\% in easy, 2.5\% in hard), suggesting
that it has learned the shape properties needed for the task.  The full system
does best and substantially better than any baseline (by 3.4\% in easy, 6.9\%
in hard). This is to be expected since Equation \ref{eqn:lossEmb},
modulo a margin parameter, aims to ensure that any positive pair is closer than any negative pair,
which is equivalent to the AUROC \cite{Fawcett06}.
Relative performance compared to the initialization consistently {\it improves} for both the full system and the attribute-only system 
when going from Easy to Hard settings, providing further evidence that the system is indeed modeling 3D properties.

\begin{table}
\centering
\caption{AUROC for the mental-rotation task. Both variants of our approach substantially out-perform the baselines.}
\label{tab:rotation}
{\small
\begin{tabular}{l@{~~~}c@{~~}c@{~~}c@{~~~}c@{~~~~~}c@{~~~~~}c@{~~}c@{~~~~~}c@{~~}c} \toprule
            & \multicolumn{3}{c}{Full Network} & Attr. Only & Pretr. & \multicolumn{2}{c}{IFV} \\
            & Emb. & FC7  & Attr & FC7 & FC7 & \cite{Lowe2004} & \cite{Arandjelovic11} \\ \midrule

All         & \bf 92.3 & 90.7 & 81.9 & 89.8 & 88.9 & 78.0 & 74.4 \\
Hard        & \bf 86.9 & 84.1 & 76.4 & 82.5 & 80.0 & 57.3 & 61.9 \\
\bottomrule
\end{tabular} 
}
\end{table}

\begin{figure}[t]
\includegraphics[width=\linewidth]{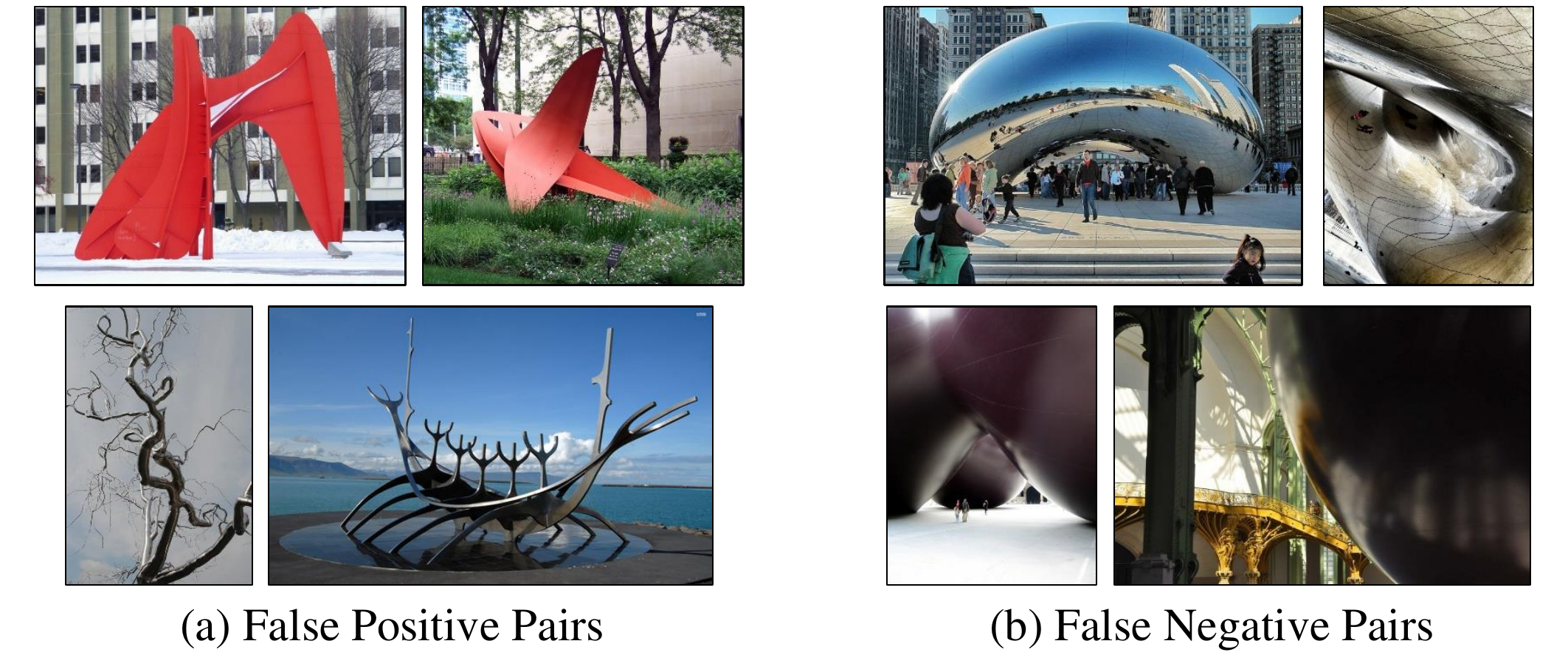}
\caption{Examples of typical mental rotation failures. Every pair in (a)
has lower distance in the embedding space than every pair in (b).}
\label{fig:rotationfail}
\end{figure}

\noindent {\bf Failure Modes:} Examining incorrect pairs reveals a number of
failure modes that suggest room for further improvement by future work.
We define mistakes by converting distances in shape embedding space
into classifications by thresholding at the equal error rate point. Figure 
\ref{fig:rotationfail} shows a few illustrative examples of these embedding
mistakes; all of the false negatives (i.e., two views of the same sculptures
that have high distance) depicted are further apart than all the false
positives (i.e., two distinct sculptures that have low distance).

The two most frequent causes of false negative pairs are specular objects
that reflect their surroundings and enormous objects that
lend themselves to being photographed from a variety of different viewpoints.
The most confused object, Anish Kapoor's {\it Cloud Gate} (`The Bean') (Fig.~\ref{fig:rotationfail}(b) top) 
combines both of these. The remaining mistakes are pairs with dramatic scale or viewpoint
changes, images where the sculpture is not the salient object, and a handful of
labeling errors.

False positive pairs tend to be works by the same artist that 
or that are similar in terms of properties.
The within-artist mistakes tended to be caused by a series of
works with a common material and theme, for instance
Alexander Calder's works with red sheet metal (e.g., Fig.~\ref{fig:rotationfail}(a) top).
Across artists, the network sometimes had difficulty
distinguishing different sculptures made with thin metal structures and between
statues of people.

\noindent {\bf Updated metadata:} The above mental rotation experiments are
done using the metadata from our prior work \cite{Fouhey16}. We have
since updated the metadata and will release the updates.  First, we
manually identified works of art in the test set that are of similar shape (i.e.,
sharing the same attributes) but exactly the same subject (e.g., busts of animals
from Ai Weiwei's {\it Zodiac Heads}) and excluded them from mental rotation
evaluation. We then trained a CNN on the entire dataset to discriminate between
works. Confident prediction mistakes ($866$) were examined for reassignment 
frequently confused sculptures ($10$) were examined for merging or exclusion.
In total $7$ works and $699$ images were updated.
Evaluating on this cleaner data leads to an increase in AUROC of about $0.1\%$;
the influence of these updates is limited since the metric is computed over
{\it pairs}: the updates affect a small fraction of images and an even smaller
number of pairs. 

\subsection{Object Characterization}
\label{sec:exp_pascal}

Our evaluation has so far focused on sculptures, and one concern is that 
what we learn may not generalize to more everyday objects like trains or cats.
We thus investigate our model's beliefs about these objects
by analyzing its activations on the PASCAL VOC dataset \cite{Everingham10}. 
We feed the windows of the trainval set of VOC-2010 
to our shape attribute model, excluding
difficult and too-small ($<100$px) windows,
and obtain a prediction of the
probability of each attribute. We probe the representation by
sorting class members by their activations (i.e., ``which trains are planar?'')
and sorting the classes by their mean activations.

\begin{figure}[t]
\includegraphics[width=\linewidth]{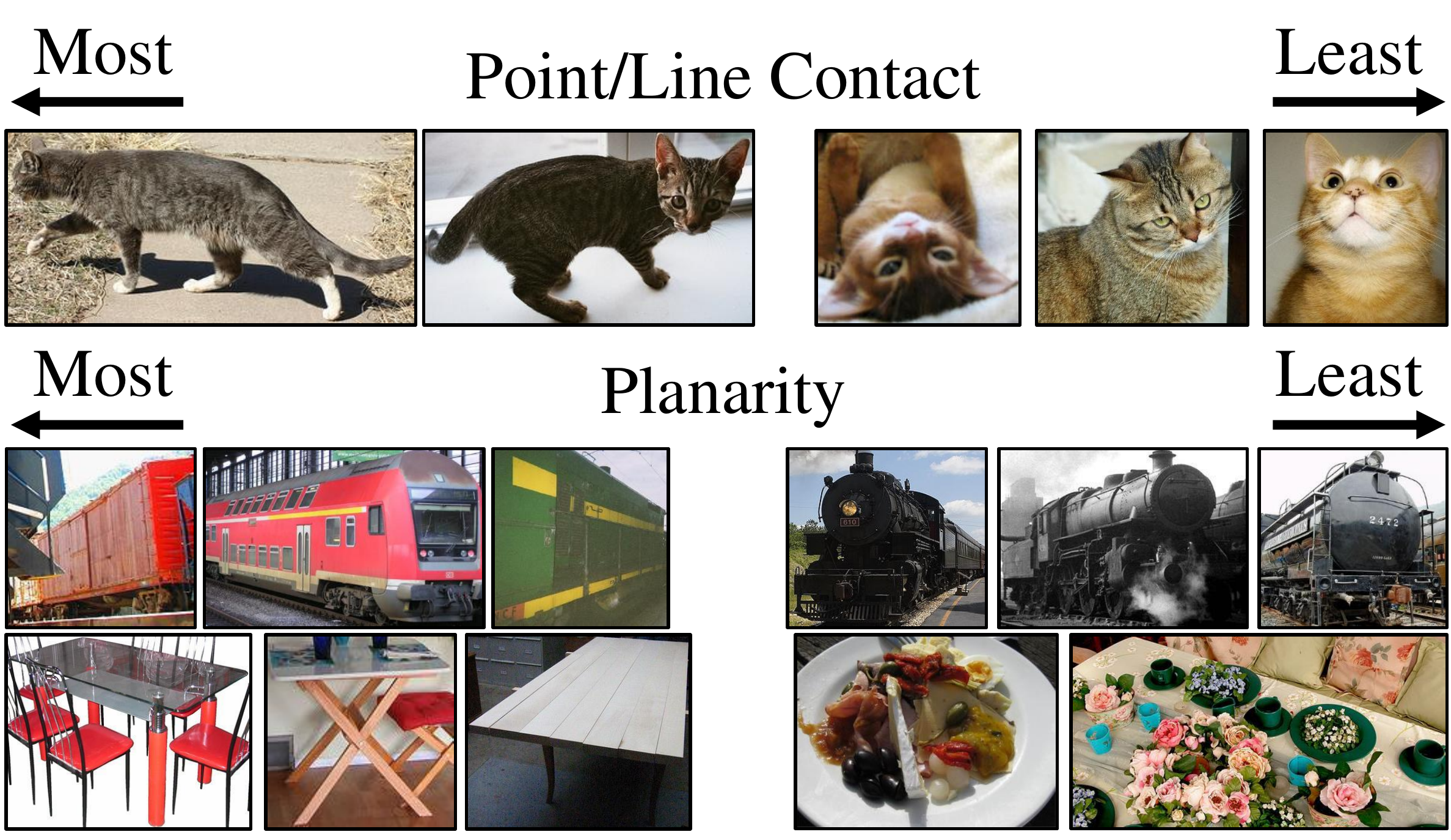} 
\caption{The top activations on PASCAL objects for Planarity and Point/Line Contact.}
\label{fig:PASCAL}
\end{figure}

\noindent {\bf Per-image results:} 
The system forms sensible beliefs about the PASCAL objects, as we show
in Fig.~\ref{fig:PASCAL}. Looking at intra-class activations,
cats lying down are predicted to have single, non-point contact as compared 
to ones standing up; trains are generally planar, except for older
cylindrical steam engines. Similarly, the non-planar dining tables are the result of occlusion by non-planar objects.

\noindent {\bf Per-category results:}
The system performs well at a category-level as well.  Note that averaging over
windows characterizes how objects {\it appear} in PASCAL VOC, not how they
are prototypically imagined: e.g., as seen in Fig.~\ref{fig:PASCAL}, the cats and
dogs of PASCAL are frequently lying down or truncated. The top 3
categories by planarity are bus, TV Monitor, train; and the bottom 3 are cow,
horse, sheep.  For point/line contact: bus, aeroplane, car are at the top
and cat, bottle, sofa are at the bottom. Finally, sheep, bird, and potted
plant are the roughest categories in PASCAL and car, bus, and aeroplane the
smoothest. 

\noindent {\bf Discriminating between classes:}
It ought to be possible to distinguish between the VOC categories based on their 
3D properties, and thus we verify that the predicted 3D shape attributes carry
class-discriminative information. We represent each window with its 12
attribute probabilities and train a random forest classifier for two outcomes 
in a 10-fold cross-validation setting: a 20-way multiclass model and a one-vs-rest. The 
multiclass model achieves an accuracy of 65\%, substantially above chance. The 
one-vs-rest model achieves an average AUROC of 89\%, with vehicles performing best.

\section{Summary and Extensions}
\label{sec:extensions}

We have shown that 3D shape attributes can be inferred directly from images at
quite high quality. In the process, we have introduced a large dataset of
modern sculpture for analyzing 3D shape attributes, verified that 
our learned models are actually inferring the attributes and not a proxy
property and analyzed what cues are being used to infer these attributes. 

One application is to use the attributes to help constrain
metric reconstruction. There has been considerable work recently
on using categories to constrain or regularize reconstruction \cite{Kar15,Ladicky11,Hane14} -- for
example roads and walls should be planar but people should not be -- and
3D shape attributes can be used similarly. In contrast to categories,
though, attributes offer a number of advantages: they can handle unseen
categories, or the open world problem; they enable sharing across categories during
learning; and they handle exceptions more easily -- some walls and 
many roads are not, in fact, planar.

Another area of investigation is extending our shape attributes -- for example, we did not
consider changes in curvature, or the presence or absence of concavities.  However, more generally,
the attributes can be extended beyond absolute and global properties.
Instead of absolute properties, many of our attributes (e.g., roughness) are
better modeled as {\em relative} attributes. An alternative is to parse objects
both globally as well as locally. For instance one could describe a sculpture
as being primarily rough, but also localize
any  small smooth regions.

\ifCLASSOPTIONcompsoc
  
  \section*{Acknowledgments}
\else
  
  \section*{Acknowledgment}
\fi
Financial support for this and our previous work was provided by the EPSRC
Programme Grant Seebibyte EP/M013774/1, ONR MURI N000141612007, Intel/NSF
Visual and Experiential Computing award IIS-1539099 and a NDSEG fellowship to
David Fouhey.  The authors thank: Olivia Wiles for tools for dataset cleaning;
Omkar Parkhi, Xiaolong Wang, and Phillip Isola for helpful conversations; and
NVIDIA for GPU donations.

\ifCLASSOPTIONcaptionsoff
  \newpage
\fi

\bibliographystyle{ieee}
\bibliography{local}

\begin{IEEEbiographynophoto}{David F. Fouhey}
is a Postdoctoral Fellow at the Electrical Engineering and Computer Science Department,
University of California, Berkeley. 
\end{IEEEbiographynophoto}

\begin{IEEEbiographynophoto}{Abhinav Gupta}
is an Assistant Professor at the Robotics Institute, Carnegie Mellon University.
\end{IEEEbiographynophoto}

\begin{IEEEbiographynophoto}{Andrew Zisserman}
is the Professor of Computer Vision Engineering at
the Department of Engineering Science, University of Oxford.
\end{IEEEbiographynophoto}

\end{document}